\documentclass[letterpaper,journal]{IEEEtran}
\usepackage{amsmath,amsfonts}
\usepackage{algorithmic}
\usepackage{algorithm}
\usepackage{array}
\usepackage[caption=false,font=normalsize,labelfont=sf,textfont=sf]{subfig}
\usepackage{textcomp}
\usepackage{stfloats}
\usepackage{url}
\usepackage{verbatim}
\usepackage{graphicx}
\usepackage{threeparttable}
\usepackage{cite}
\usepackage{booktabs}
\usepackage{enumitem}
\usepackage{placeins} 
\usepackage{multicol}
\usepackage{multirow}
\usepackage{float}
\usepackage{siunitx}

\usepackage{tabularx}
\usepackage{ragged2e}
\usepackage{hyperref}
\usepackage{amssymb}
\usepackage{tabularx}

\usepackage{graphicx}      
\usepackage{subcaption}    
\usepackage{amsmath}       
\DeclareUnicodeCharacter{2248}{\approx}

\begin{document}
\title{DGTEN: A Robust Deep Gaussian based Graph Neural Network for Dynamic Trust Evaluation with Uncertainty-Quantification Support}

\author{
\IEEEauthorblockN{Muhammad Usman\IEEEauthorrefmark{1}\thanks{Corresponding author: Muhammad Usman (mucbp@umkc.edu)} 
and Yugyung Lee\IEEEauthorrefmark{1}}
\\
\IEEEauthorblockA{\IEEEauthorrefmark{1}Department of Computer Science,
University of Missouri-Kansas City,\\
5100 Rockhill Road, Kansas City, MO 64110, USA\\
Email: \{mucbp, leeyu\}@umkc.edu}
}

\markboth{Journal of \LaTeX\ Class Files,~Vol.~\#, No.~\#, August~2025}%
{Usman \MakeLowercase{\textit{et al.}}: DGTEN: A Robust Deep Gaussian based Graph Neural Network for Dynamic Trust Evaluation with Uncertainty-Quantification Support}

\IEEEpubid{\makebox[\columnwidth]{0000--0000/00/\$00.00~\copyright~2025 IEEE\hfill}%
\hspace{\columnsep}\makebox[\columnwidth]{}}

\maketitle

\begin{abstract}
Dynamic trust evaluation in large, rapidly evolving graphs demands models that capture changing relationships, express calibrated confidence, and resist adversarial manipulation. DGTEN (Deep Gaussian based Trust Evaluation Network) introduces a unified graph-based framework that does all three by combining uncertainty-aware message passing, expressive temporal modeling, and built-in defenses against trust-targeted attacks. It represents nodes and edges as Gaussian distributions so that both semantic signals and epistemic uncertainty propagate through the graph neural network, enabling risk-aware trust decisions rather than overconfident guesses. To track how trust evolves, it layers hybrid absolute–Gaussian–hourglass positional encoding with Kolmogorov–Arnold network based unbiased multi-head attention, then applies an ordinary differential equation–based residual learning module to jointly model abrupt shifts and smooth trends. Robust adaptive ensemble coefficient analysis prunes or down-weights suspicious interactions using complementary cosine and Jaccard similarity, curbing reputation laundering, sabotage, and on–off attacks. On two signed Bitcoin trust networks, DGTEN delivers standout gains where it matters most: in single-timeslot prediction on Bitcoin-OTC, it improves MCC by +12.34\% over the best dynamic baseline; in the cold-start scenario on Bitcoin-Alpha, it achieves a +25.00\% MCC improvement, the largest across all tasks and datasets while under adversarial on–off attacks it surpasses the baseline by up to 10.23\% MCC. These results endorse the unified DGTEN framework.
\end{abstract}



\section{Introduction}

\IEEEPARstart{T}{rust} is the belief or confidence one entity places in another within a specific context, serving to mitigate the risks inherent in interactions and communications. It is inherently \emph{subjective} (varies between individuals), \emph{dynamic} (evolves over time), context-dependent, asymmetric (directional and non-reciprocal), and exhibits conditional transferability and composability.

In computational settings, trust evaluation quantifies the degree to which a trustor (entity placing trust) believes in a trustee (entity being trusted), often with respect to security, usability, maintainability, and reliability. When machine learning methods are employed to infer future trust relationships from historical interaction data, the task is referred to as trust prediction or trust evaluation.

Modern digital ecosystems ranging from IoT deployments to social platforms, financial systems, and collaborative networks are characterized by unprecedented interconnectivity~\cite{chen2014trust, wang2020survey, yan2014survey}. This connectivity enables transformative services but also exposes systems to sophisticated cyber threats that can undermine operational integrity~\cite{10925363,10.1145/3460120.3484805}. In this context, trust evaluation is a fundamental mechanism for systematically assessing entity reliability in networked systems~\cite{yan2014survey, wang2024trustguard}.

Trust differs from static security measures in that it is shaped by ongoing interactions, behavioral observations, and temporal patterns~\cite{10925363, wang2024trustguard}. Its core operational properties include asymmetry, propagation through intermediaries, and temporal decay, whereby recent interactions carry greater weight~\cite{lin2021medley, wang2024trustguard}. Neglecting these properties can result in undetected breaches, misinformation spread, and cascading systemic failures~\cite{9936655, JAFARIAN2025125391}.

Graph Neural Networks (GNNs)~\cite{4700287} provide a natural paradigm for modeling trust relationships as graphs, enabling end-to-end learning via message passing~\cite{yu2023kgtrust, gao2020gatrust, JAFARIAN2025125391}. Early methods such as Guardian~\cite{li2020guardian} applied GCN-based trust propagation, while GATrust~\cite{gao2020gatrust} incorporated attention-based multi-aspect attributes. Later, TrustGNN~\cite{huo2023trustgnn} modeled trust chains, and TrustGuard~\cite{wang2024trustguard} integrated temporal dynamics with basic robustness measures. However, existing approaches face three persistent limitations:

\vspace{0.5\baselineskip}

\noindent \textit{Gap 1 Inadequate dynamic modeling with uncertainty quantification:}
Many models omit temporal dynamics or rely on oversimplified discrete encodings. TrustGuard~\cite{wang2024trustguard} models time but lacks principled uncertainty estimation; Medley~\cite{lin2021medley} depends on fine-grained timestamps—often unavailable—and also lacks uncertainty modeling. No prior work jointly models continuous trust evolution and uncertainty, a critical requirement for risk-aware cybersecurity decisions.

\vspace{0.5\baselineskip}

\noindent \textit{Gap 2: Limited robustness against sophisticated dynamic attacks:}
Trust systems remain vulnerable to manipulations such as bad-mouthing, good-mouthing, and on--off attacks. Existing defenses (e.g., similarity-based pruning) fail to adapt to coordinated or evolving adversarial strategies~\cite{10.1145/3460120.3484805, wang2024trustguard}.

\vspace{0.5\baselineskip}

\noindent \textit{Gap 3: Lack of integrated architectures:}
Temporal modeling, uncertainty quantification, and robustness mechanisms are often treated as isolated components, leading to suboptimal performance under complex or adversarial conditions.

To address these challenges, we propose DGTEN, a GNN-based architecture designed to bridge these gaps through the following components:

\subsubsection{\textbf{Deep Gaussian Message Passing (DGMP)}}
DGMP is a Graph Convolutional based node embedding mechanism that enables explicit uncertainty estimation. It maps nodes and edges to Gaussian distributions and propagates them through message passing, eliminating the need for post-hoc calibration.

\subsubsection{\textbf{RAECA Defense}}
Robust Adaptive Ensemble Coefficient Analysis (RAECA) uses cosine and Jaccard similarities to identify and mitigate trust-related attacks.

\subsubsection{\textbf{Adaptive Temporal Framework}}
This temporal framework models how a node's trust level evolves over time and consists of:

\begin{enumerate}[label=\alph*)]
\item \textit{\textbf{Hybrid Absolute Gaussian Hourglass (HAGH):}} A positional encoding that assigns intuitive and distinguishable temporal identities to node embeddings, improving further temporal pattern modeling.

\item \textit{\textbf{KAN Layer-Based Multi-Head Attention:}} To capture nonlinear and unbiased dependencies in dynamic trust, we integrate a Chebyshev polynomial-based Kolmogorov-Arnold Network (KAN) layer into the multi-head self-attention mechanism.

\item \textit{\textbf{Neural ODE-Based Residual Learning:}} Neural Ordinary Differential Equation (ODE)-based residual learning is introduced to model continuous-time trust evolution.
\end{enumerate}

With these contributions DGTEN outperforms state-of-the-art methods by up to $+12.34\%$ in MCC for single-timeslot prediction and $+25.00\%$ in cold-start scenarios, with consistent AUC, balanced accuracy, and F1-score gains under adversarial conditions on dynamic Bitcoin datasets. The remainder of this paper is organized as follows: Section~\ref{sec:related} reviews related literature; Section~\ref{sec:method} presents the problem formulation and the DGTEN architecture; Section~\ref{sec:experiments} describes the experimental design and results; Section~\ref{sec:discussion} discusses implications and practical considerations; and Section~\ref{sec:conclusion} concludes with key insights and future research directions.

\section{Related Work}
\label{sec:related}

Trust evaluation underpins the security of cyber-physical and information systems by enabling entities to assess the reliability of peers despite dynamic behaviors and potential adversarial manipulation~\cite{wang2020survey}. From a machine learning perspective, trust evaluation is fundamentally a representation learning problem on dynamic, signed, and often adversarial graphs. Existing approaches can be broadly categorized into  {statistical},  {reasoning-based}, and  {machine learning} paradigms~\cite{wang2024trustguard}. Statistical models aggregate interaction priors using frequentist measures. While computationally efficient, they suffer from the cold-start problem due to their reliance on dense historical support, making them poor approximators for sparse or emerging nodes~\cite{wang2024trustguard}. Reasoning-based methods, such as those grounded in Subjective Logic, rely on hard-coded axiomatic rules for trust propagation. These methods introduce strong inductive biases that often fail to generalize in heterogeneous environments where trust transitivity is non-linear or context-dependent~\cite{wang2020survey}. Consequently, Graph Neural Networks (GNNs)\cite{4700287} have emerged as the dominant paradigm, treating trust evaluation as a link prediction or node regression task within an end-to-end message-passing framework~\cite{suarez2022graph,li2024permutation,huo2023trustgnn}.

Early GNN-based architectures focused on encoding static snapshots of trust networks. GCN-based models typically employ isotropic aggregation, propagating features uniformly from neighbors~\cite{li2020guardian,zhan2024enhancing,jiang2024tfd,wang2024joint}. From a representation learning standpoint, this isotropic inductive bias is suboptimal for trust graphs, as it fails to differentiate between high-reliability and low-reliability signals during feature aggregation. To address this, attention-based mechanisms (GATs)\cite{vaswani2017attention} were introduced to learn anisotropic weights, allowing the model to attend differentially to neighbors based on node or edge features~\cite{gao2020gatrust,yu2023kgtrust,bellaj2023gbtrust}. While this improves expressivity, the quadratic complexity of self-attention relative to node degree often hinders scalability. Alternative approaches explicitly model trust paths as distinct compositional chains~\cite{huo2023trustgnn}, injecting structural priors about transitivity. However, these static architectures fundamentally ignore the temporal dimension, failing to capture non-stationary dynamics such as reputation decay or sudden behavioral drifts.

Temporal GNNs have been proposed to better capture non-stationary trust dynamics by moving from static to time-aware graph modeling. In discrete-time designs, a static GNN encoder is usually combined with recurrent units (such as GRUs) or temporal attention mechanisms to handle sequences of graph snapshots~\cite{wang2024trustguard,wen2023dtrust,JAFARIAN2025125391}. However, these approaches face a fundamental snapshot granularity trade-off: using coarse snapshots obscures fast, adversarial behaviors (e.g., rapid on–off attacks), while using fine-grained snapshots produces very sparse graphs and can cause vanishing gradients over long temporal sequences. Continuous-time dynamic graph networks (CTDGNs) address this by encoding timestamps directly into the embedding space, preserving fine-grained temporal fidelity~\cite{lin2021medley}. However, encoding continuous time often requires complex functional embeddings or point-process modeling, resulting in high inference latency that is impractical for large-scale, real-time systems. Furthermore, while some temporal models incorporate adversarial defenses such as cosine-similarity-based edge pruning, these defenses are typically heuristic and decoupled from the learning objective, limiting their robustness against adaptive gradient-based attacks~\cite{wang2024trustguard,JAFARIAN2025125391}.

Despite these advancements, several fundamental learning challenges remain unresolved in the current state-of-the-art. First, existing GNNs for trust are predominantly deterministic, producing point estimates that fail to capture epistemic uncertainty. In safety-critical and adversarial settings, the inability to distinguish between aleatoric noise and epistemic ignorance (due to out-of-distribution data or sparsity) leads to overconfident predictions on unreliable nodes. Second, the integration of temporal dynamics remains rigid; models either struggle with long-term dependencies in discrete sequences or face scalability bottlenecks in continuous domains, lacking a hybrid mechanism to efficiently model both smooth trends and abrupt shifts. Third, robustness mechanisms are rarely end-to-end differentiable or adaptive. Most defenses rely on fixed similarity thresholds that can be easily circumvented by adversarial perturbations designed to mimic homophily. Finally, these components, uncertainty quantification, temporal dynamics, and robustness are typically treated as orthogonal modules rather than being unified into a cohesive probabilistic learning framework~\cite{wang2024trustguard,lin2021medley,wen2023dtrust}. There is a clear need for architectures that inherently model uncertainty within the message-passing phase and leverage this uncertainty for node related decision making. 


\begin{table*}[!t]
\centering
\small
\caption{Comparative Analysis of state of the art Trust Evaluation Models}
\label{tab:model_comparison}
\begin{tabular}{@{}l l c c l@{}}
\toprule
\textbf{Model} & \textbf{Approach/Domain} & \textbf{Uncert.} & \textbf{Robust.} & \textbf{Key Limitations} \\
\midrule
\multicolumn{5}{@{}l}{\textit{\textbf{Static Models}}} \\
Guardian\cite{li2020guardian}      & GC/OSN            & --     & --      & No temporal dynamics; ignores node attributes. \\
TREF\cite{zhan2024enhancing}       & GC+EK/MC          & --     & --      & Domain-specific; no temporal modeling. \\
T-FrauDet\cite{jiang2024tfd}       & ANN+TAF/SIoT      & --     & Partial & Classification-oriented; static core. \\
JoRTGNN\cite{wang2024joint}        & Hetero-GC/OSN     & --     & --      & Task-specific; poor scalability. \\
GATrust\cite{gao2020gatrust}       & GAT/OSN           & --     & --      & Static; scalability concerns. \\
GBTrust\cite{bellaj2023gbtrust}    & GAT-E/P2P         & --     & Partial & P2P-specific; no uncertainty handling. \\
KGTrust\cite{yu2023kgtrust}        & HetAtt+KG/SIoT    & --     & --      & Requires external KG; no temporal modeling. \\
TrustGNN\cite{huo2023trustgnn}     & Chain-based/OSN   & --     & --      & Hyperparameter-sensitive; limited generalization. \\
\midrule
\multicolumn{5}{@{}l}{\textit{\textbf{Discrete-Time Models}}} \\
DTrust\cite{wen2023dtrust}         & GCN+GRU/OSN       & --     & --      & Misses fine-grained changes; scalability issues. \\
MATA\cite{JAFARIAN2025125391}      & GCN+GRU+Att/OSN   & --     & Partial & Reputation module requires manual tuning. \\
TrustGuard\cite{wang2024trustguard}& PAA/Generic       & --     & Full    & Relies heavily on homophily assumption. \\
\midrule
\multicolumn{5}{@{}l}{\textit{\textbf{Continuous-Time Models}}} \\
Medley\cite{lin2021medley}         & Cont. Att/OSN     & --     & --      & Needs fine timestamps; high overhead. \\
\midrule
\multicolumn{5}{@{}l}{\textit{\textbf{Discrete+Continues-Time Models}}} \\
DGTEN                         & DGMP+ODE/Generic & DGMP  & Full    & Potential Complex Architecture. \\
\bottomrule
\end{tabular}

\vspace{0.3em}
\footnotesize
\textbf{Abbreviations:} GC: Graph Convolution; EK: Expert Knowledge; TAF: Time-Aggregated Features; 
GAT: Graph Attention; GAT-E: GAT Edge-level; HetAtt: Heterogeneous Attention; 
KG: Knowledge Graph; GRU: Gated Recurrent Unit; PAA: Position-aware Attention; 
DGMP: Deep Gaussian Message Passing; OSN: Online Social Network; 
MC: Mobile Crowdsourcing; SIoT: Social IoT; P2P: Peer-to-Peer.
\vspace{-3mm}
\end{table*}

\section{Methods and Materials}
\label{sec:method}
This section presents the problem definition, mathematical formulation, and architectural design of the DGTEN model for node uncertainty-aware dynamic trust evaluation.
\vspace{-2mm}

\subsection{Problem Definition and Formulation}\label{subsec:problem_definition_formulation}
We address trust evaluation on a dynamic graph $G$ that evolves during an observation period ending at $T_{\text{obs}}$. During this period, the active node set $V(t) \subseteq V_{\text{global}}$ and edge set $E(t)$ change over time $t \in [0, T_{\text{obs}}]$, where $V_{\text{global}}$ denotes all unique nodes in the system.

We discretized the dynamic graph into $N$ ordered snapshots $\{G_1, G_2, \dots, G_N\}$, each aggregating activity over discrete intervals of length $\Delta t_{\text{snap}}$. The $k^{\text{th}}$ snapshot $G_k = (V_k, E_k)$ captures interactions during $[(k-1)\Delta t_{\text{snap}}, k\Delta t_{\text{snap}}]$, where $V_k \subseteq V_{\text{global}}$ denotes active nodes and $E_k$ contains weighted, directed edges between them. The $k^{\text{th}}$ snapshot ends at $t_k = k\Delta t_{\text{snap}}$, with the observation period concluding at $t_N = T_{\text{obs}}$.

An edge $e_{i \rightarrow j}^{(w,k)} \in E_k$ from node $i \in V_k$ to $j \in V_k$ with rating $w$ indicates that $i$ trusts $j$ at level $w$ during timeslot $k$. The trust level $w$ is a categorical label from set $\mathcal{W}$ (e.g., $\mathcal{W} = \{\text{'Distrust', 'Trust'}\}$ for binary trust), with granularity varying by application.

We define the trust evaluation problem as developing a model $\text{DGTEN}(\cdot)$ with dual objectives: (1) predicting trust level $w_{\text{pred}} \in \mathcal{W}$ for an edge from source $i$ to target $j$ in a future timeslot (typically of duration $\Delta t_{\text{snap}}$ after the observation period, e.g., $(t_N, t_N + \Delta t_{\text{snap}}]$), and (2) learning the node uncertainties in the observation period. Predictions concern node pairs from $V_{\text{global}}$, particularly those active through snapshot $G_N$. A key challenge is achieving both objectives accurately despite potential local attacks (anomalies, deceptions, or malicious behaviors) in the historical data up to $t_N$.
\vspace{-2mm}
\subsection{Trust-Related Attacks and Their Simulation}
Trust-related attacks manipulate reputation in dynamic graphs through dishonest ratings or strategic behavior. We simulate four primary attack strategies where malicious nodes artificially inflate or undermine reputations.

In a \textit{good-mouthing} (ballot-stuffing) attack, malicious nodes inflate positive ratings to boost their own or collaborators' trust scores. We simulate this by selecting a random 10\% subset of untrustworthy 'bad nodes' (with more incoming distrust than trust) as victims. For each victim, we choose attackers nodes equal to its total degree, prioritizing those at maximum shortest-path distance, and have each attacker add one new trust edge to the victim. This scales the attack with victim connectivity while avoiding duplicate edges.

In a \textit{bad-mouthing} attack, malicious nodes issue false negative ratings to damage well-behaved nodes' reputations. Our simulation mirrors the good-mouthing approach but targets reliable 'good nodes' (with more incoming trust than distrust connections). A random 10\% of these nodes are selected as victims, and new distrust edges matching the victim's degree are added from most distant attackers node to tarnish their reputations.

In an \textit{on-off} (conflict behavior) attack, malicious nodes alternate between honest and dishonest actions over time to evade detection while maintaining positive reputation. This temporal inconsistency misleads trust models, particularly those reliant on short-term patterns, causing overestimated trustworthiness~\cite{wang2020survey, chen2016trust, chen2014trust}. We simulate this by intermittently applying the bad-mouthing attack across snapshots: during an \textit{on-phase} at time $t$, a full bad-mouthing attack executes on 10\% of good nodes, while in the subsequent \textit{off-phase} at $t+1$, no malicious edges are added, allowing potential reputation recovery. This alternation mimics natural fluctuations, enabling long-term trust erosion while evading detection.

In a \textit{slow-poisoning} attack, malicious nodes gradually expand a good-mouthing campaign over time so that the attack effect in the trust snapshots builds up slowly rather than appearing as a sudden anomaly. The set of victim nodes is not fixed once at the start; instead, at each snapshot \(t\) in a sequence of \(T\), we randomly select up to \(0.15 \cdot (t/T)\) of structurally bad nodes (nodes with more incoming distrust than trust) to be attacked. For every selected victim, we reuse the same edge-injection procedure as in the good-mouthing attack, choosing attacker nodes in proportion to the victim's total degree and prioritizing those that are farthest away in terms of shortest-path distance, with each attacker adding a single new trust edge. As \(t\) increases, both the proportion of victim nodes and the number of fabricated trust edges grow linearly, leading to a gradual and hard-to-detect shift in overall reputation rather than a sharp disruption

We use these trust related-attacks as it form the basis of most adversarial behaviors in trust networks and can be used to simulate a wide range of trust-manipulation strategies. They alter or introduce synthetic edges (interactions), enabling stealthy yet effective disruption of trust inference~\cite{Jin_Feng_Guo_Wang_Wei_Wang_2023,10.1109/TKDE.2022.3201243}. Coordinated collusion among multiple attackers is substantially more damaging than isolated actions~\cite{chen2014trust}, and systematic dishonest ratings erode model reliability~\cite{jagielski2021subpopulation}. Mitigating these fundamental attack patterns is critical, as countering them effectively blocks most higher-level variants.

\begin{figure}[!t]
    \centering
    \includegraphics[width=0.89\columnwidth]{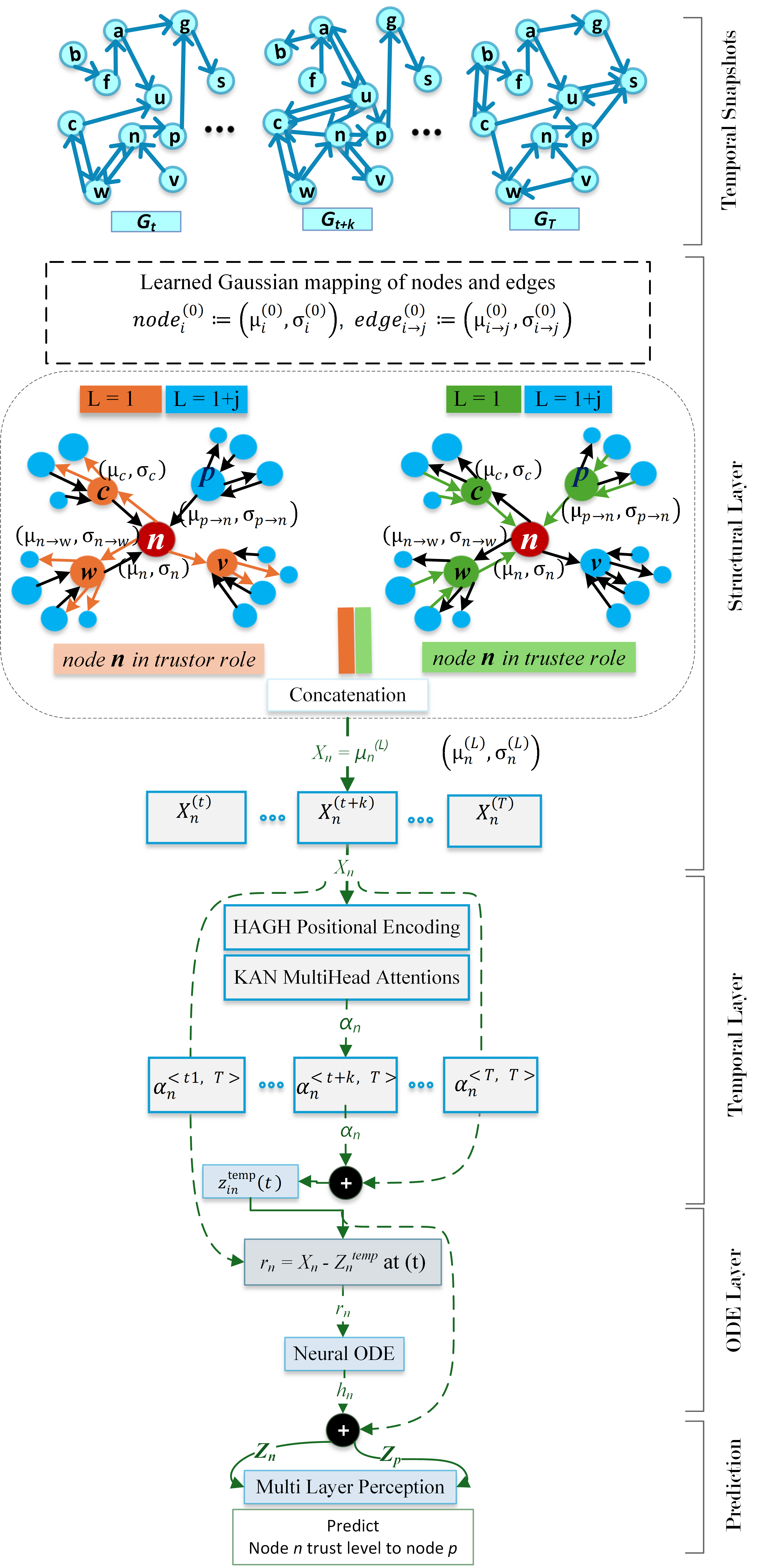}
    \caption{The architecture of DGTEN.}
    \label{fig:dgten_arc}
    \vspace{-5mm}
\end{figure}
\vspace{-3.3mm}
\subsection{The Architecture of the DGTEN Model}
The architecture of DGTEN comprises three interconnected sub-models, each responsible for a distinct yet complementary aspect of trust modeling: (1) the \textit{Structural GNN Model}, (2) the \textit{Temporal GNN Model}, and (3) \textit{ODE Residual Learning}. The structural component generates node-level embeddings (NEs) from input graph snapshots using our introduced Deep Gaussian-based graph convolution mechanism, which captures trust spatial relationships along with uncertainty-aware message passing and aggregation. These NEs are stacked snapshot-wise temporally and further modeled by the temporal layer to capture the dynamic evolution of trust over time. The temporal modeling begins with HAGH Positional Encoding, enriching NEs with chronological context. This is followed by a multi-head self-attention mechanism that leverages Chebyshev-KANs for expressive, nonlinear transformations of temporally-aware embeddings. Finally, an ODE-based residual learning mechanism refines the temporal trajectories, allowing the model to capture both discrete jumps and smooth transitions in trust dynamics.

\subsection{The Structural Layer (SL)}
\label{sec:structural_layer_overall}
The SL of DGTEN learns NEs) and associated node uncertainty vectors by modeling spatial dependencies in the trust graph with a Deep Gaussian Message Passing architecture. It captures local and multi-hop trust relationships and produces role-aware representations by separately modeling the trustor (interaction initiator) and trustee (interaction receiver) roles before integrating them into a unified node representation. To address adversarial interactions, the layer employs RAECA, which scores and prunes suspicious edges using Jaccard and cosine similarity. The layer also estimates node-level uncertainty, supporting risk-aware decision making and distinguishing confident from uncertain trust assessments.

\subsubsection{Gaussian Input Representation}
\label{sec:gaussian_input_rep}

Before propagation begins, we map deterministic inputs into the probabilistic Gaussian space.

\paragraph{Node Mapping to Gaussian}
\label{sec:node_mapping}

Each node $i \in \mathcal{V}$ is initialized with a Gaussian representation
\[
\mathbf{h}_i^{(0)} := \left(\boldsymbol{\mu}_i^{(0)}, \boldsymbol{\sigma}_i^{(0)}\right) \in \mathbb{R}^{d'},
\]
where $\boldsymbol{\mu}_i^{(0)}$ encodes semantic content and $\boldsymbol{\sigma}_i^{(0)}$ quantifies feature-wise uncertainty. Raw node features $\mathbf{x}_i \in \mathbb{R}^f$ (application-specific attributes, random initializations, or pre-trained embeddings such as Node2Vec \cite{grover2016node2vec}) are first projected to a shared latent space of dimension $d'$:
\[
\mathbf{x}_i^{\text{input}} = \mathbf{W}^{(\text{proj})} \mathbf{x}_i + \mathbf{b}^{(\text{proj})}.
\]
A sinusoidal mapping, inspired by random Fourier features, enriches this representation:
\[
\mathbf{p}_i = \mathbf{W}^{(\text{freq})}\mathbf{x}_i^{\text{input}} + \mathbf{b}^{(\text{freq})}, \quad
\mathbf{x}_i^{\text{rff}} = \left[\cos\left(\mathbf{p}_i\right), \sin\left(\mathbf{p}_i\right)\right] \in \mathbb{R}^{d'}.
\]
This introduces nonlinearity and periodicity for more expressive encodings. The initial Gaussian parameters are then obtained via parallel linear projections:
\[
\boldsymbol{\mu}_i^{(0)} = \mathbf{W}_{\mu} \mathbf{x}_i^{\text{rff}} + \mathbf{b}_{\mu}, \qquad
\mathbf{s}_i^{(0)} = \mathbf{W}_{\sigma} \mathbf{x}_i^{\text{rff}} + \mathbf{b}_{\sigma}.
\]
The standard deviation is given by
\[
\boldsymbol{\sigma}_i^{(0)} \leftarrow \max(\exp(0.5 \cdot \log \mathbf{s}_i^{(0)}), \sigma_{\min}),
\]
where $\sigma_{\min}$ is a small constant for numerical stability. This logarithmic parameterization guarantees strictly positive variances and stabilizes training.

\paragraph{Edge Label Gaussian Mapping}
\label{sec:edge_label_gaussian}

At each graph convolution layer $k$, raw edge labels $\ell^{\text{raw}}_{i\to j}$ are mapped to Gaussian edge embeddings $\mathbf{e}^{(k)}_{i\to j} = (\boldsymbol{\mu}^{(k)}_{i\to j}, \boldsymbol{\sigma}^{(k)}_{i\to j})$. This transformation is  {stateless across layers}: unlike node embeddings which evolve across layers, edge embeddings are recomputed independently at each layer directly from raw labels. This allows each layer to independently reinterpret edge semantics at different network depths, providing flexible message passing modulation. A sinusoidal mapping is applied using layer-specific parameters:
\[
\mathbf{p}_{i \rightarrow j}^{(k)} = \mathbf{W}^{(\text{edge},k)} \boldsymbol{\ell}_{i\rightarrow j}^{\text{raw}} + \mathbf{b}^{(\text{edge},k)}
\]
\[
\boldsymbol{\ell}_{i\rightarrow j}^{\text{rff},(k)} = [\cos(\mathbf{p}_{i\rightarrow j}^{(k)}), \sin(\mathbf{p}_{i\rightarrow j}^{(k)})] \in \mathbb{R}^{d_\ell},
\]
where $\mathbf{W}^{(\text{edge},k)} \in \mathbb{R}^{\frac{d_\ell}{2} \times f_\ell}$, $\mathbf{b}^{(\text{edge},k)} \in \mathbb{R}^{\frac{d_\ell}{2}}$. The Gaussian parameters are:
\[
\boldsymbol{\mu}_{i\rightarrow j}^{(k)} = \mathbf{W}_{\mu}^{(\text{edge},k)} \boldsymbol{\ell}_{i\rightarrow j}^{\text{rff},(k)} + \mathbf{b}_{\mu}^{(\text{edge},k)},
\]
\[
\mathbf{s}_{i\rightarrow j}^{(k)} = \mathbf{W}_{\sigma}^{(\text{edge},k)} \boldsymbol{\ell}_{i\rightarrow j}^{\text{rff},(k)} + \mathbf{b}_{\sigma}^{(\text{edge},k)}
\]
\[
\boldsymbol{\sigma}_{i\rightarrow j}^{(k)} \leftarrow \max(\exp(0.5 \cdot \log\mathbf{s}_{i\rightarrow j}^{(k)}), \sigma_{\min}).
\]

\subsubsection{Deep Gaussian Message Passing (DGMP)}
\label{sec:dgmp_process}

We define Deep Gaussian Message Passing (DGMP) as the core layer-wise mechanism that processes the initialized Gaussian inputs. Unlike standard message passing, DGMP explicitly integrates an adversarial defense step (RAECA) with a probabilistic aggregation step. For any layer $k$, the DGMP process consists of two sequential operations: first, computing robust reliability coefficients ($\alpha$) to filter attacks, and second, aggregating Gaussian distributions using those coefficients.

\paragraph*{Step 1: RAECA Defense}
\label{sec:raeca}

To ensure robustness against adversarial behavior such as related attacks \cite{ijcai2019p669,10.1145/3394486.3403049}, the SL employs RAECA, a pairwise association-based edge weighting mechanism grounded in network homophily theory \cite{facd5bfb-620b-3f71-8412-0c6d4aa71a2e}, trust correlation analyses \cite{chen2014trust,9546070}, and empirical attack models \cite{10.1145/2433396.2433405}. Real-world trust graphs such as Bitcoin OTC and Bitcoin Alpha exhibit strong homophily; following Zhu et al.~\cite{zhu2020beyond}, homophily ratios of 0.90 and 0.94 exceed the classical 0.7 threshold. As similar users tend to form consistent trust links \cite{chen2014trust,9546070,10.1145/2433396.2433405}, adversarial attacks commonly exploit dissimilarity by connecting targets to malicious or semantically distant neighbors \cite{ijcai2019p669,10.1145/3394486.3403049}. RAECA mitigates this by weighting edges by similarity and pruning interactions deemed unreliable.

\textit{a) Similarity computation:}
We compute two complementary similarity metrics between node $i$ and its neighbor $j$ using their embeddings from the previous layer $(k-1)$. Cosine similarity (shifted to $[0,2]$) captures directional alignment, while Fuzzy Jaccard similarity captures magnitude-aware feature overlap:
\[
S_{ij}^{\text{cos}} = 1 + \frac{\boldsymbol{\mu}_i^{(k-1)} \cdot \boldsymbol{\mu}_j^{(k-1)}}{\|\boldsymbol{\mu}_i^{(k-1)}\| \, \|\boldsymbol{\mu}_j^{(k-1)}\| + \epsilon},
\]
\[
S_{ij}^{\text{jac}} = \frac{\sum_{l=1}^{d'} \min(\mu_{i,l}^{(k-1)}, \mu_{j,l}^{(k-1)})}{\sum_{l=1}^{d'} \max(\mu_{i,l}^{(k-1)}, \mu_{j,l}^{(k-1)}) + \epsilon}.
\]
In these equations, $d'$ represents the dimensionality of the node embedding vectors, and $\epsilon$ is a small constant term added to the denominator to ensure numerical stability.

\textit{b) Similarity pruning:}
Edges falling below specific thresholds are treated as unreliable (e.g., potential attacks or noise) and are pruned using hard-thresholding:
\[
\tilde{S}_{ij}^{\text{cos}} = S_{ij}^{\text{cos}} \cdot \mathbb{I}(S_{ij}^{\text{cos}} \geq \tau^{\text{cos}}), \quad
\tilde{S}_{ij}^{\text{jac}} = S_{ij}^{\text{jac}} \cdot \mathbb{I}(S_{ij}^{\text{jac}} \geq 0.05).
\]
Here, $\mathbb{I}(\cdot)$ denotes the indicator function which returns 1 if the condition is met and 0 otherwise, effectively zeroing out contributions from dissimilar neighbors. The parameter $\tau^{\text{cos}}$ is a hyperparameter selected via grid search (see Fig.~\ref{fig:hyperparams}) to maximize F1-macro under adversarial settings.

\textit{c) Adaptive ensemble fusion:}
The valid scores are fused into a single reliability score $\tilde{S}_{ij}^{(k)}$ using a quadratic mean formulation. This approach rewards edges that have strong signals in at least one metric:
\[
\tilde{S}_{ij}^{(k)} = \frac{(\tilde{S}_{ij}^{\text{cos}})^{2} + (\tilde{S}_{ij}^{\text{jac}})^{2}}{\tilde{S}_{ij}^{\text{cos}} + \tilde{S}_{ij}^{\text{jac}} + \epsilon}.
\]
The resulting $\tilde{S}_{ij}^{(k)}$ acts as the unified raw weight for the edge $j \to i$.

\textit{d) Degree-aware normalization:}
To properly balance the influence of neighbors against the node's own self-loop, we normalize the neighbor scores relative to the pruned in-degree. This ensures that high-degree nodes do not have their own history drowned out by the sheer number of neighbors:
\[
\hat{r}_{i\leftarrow j}^{(k)} = \frac{\tilde{S}_{ij}^{(k)}}{\sum_{p \in \mathcal{N}_i^{\text{in}}} \tilde{S}_{ip}^{(k)} + \epsilon} \cdot D_i'^{(k)}.
\]
where $\mathcal{N}_i^{\text{in}}$ denotes the set of incoming neighbors for node $i$, and $D_i'^{(k)} = |\{j \in \mathcal{N}_i^{\text{in}} \mid \tilde{S}_{ij}^{(k)} > 0\}|$ represents the effective node degree after pruning unreliable edges.

\textit{e) Self-loop reinsertion \& coefficient normalization:}
Finally, a unit-weight self-loop is added to preserve node identity, and the coefficients are normalized to sum to one over the set of neighbors including the node itself $\mathcal{N}_i' = \mathcal{N}_i^{\text{in}} \cup \{i\}$:
\[
\mathbf{A}_{ij}^{(k)} =
\begin{cases}
\hat{r}_{i\leftarrow j}^{(k)} & \text{if } i \neq j, \\
1 & \text{if } i = j,
\end{cases}
\quad
\alpha_{i\leftarrow j}^{(k)} = \frac{\mathbf{A}_{ij}^{(k)}}{\sum_{p \in \mathcal{N}_i'} \mathbf{A}_{ip}^{(k)}}.
\]
The final output $\alpha_{i\leftarrow j}^{(k)}$ represents the normalized importance weight of neighbor $j$'s information for node $i$. A symmetric variant $\alpha_{j \leftarrow i}^{(k)}$ is computed similarly for outgoing edges when node $i$ acts as the trustor.

\paragraph*{Step 2: Gaussian Message Construction}
\label{sec:prob_message_construction_k}

Using these defensive coefficients $\alpha_{j \leftarrow i}^{(k)}$, node $i$ aggregates incoming and outgoing Gaussian messages separately. For incoming messages,
\[
\boldsymbol{\mu}_{\text{in},i}^{(k)}
= \sum_{j} \alpha_{i\leftarrow j}^{(k)} \, \boldsymbol{\mu}_{j}^{(k-1)},
\qquad
\boldsymbol{\sigma}_{\text{in},i}^{(k)}
= \sum_{j} \left|\alpha_{i\leftarrow j}^{(k)}\right| \, \boldsymbol{\sigma}_{j}^{(k-1)}.
\]
The edge opinion is the Gaussian embedding of an edge that represents the model's current assessment of the trust interaction on that edge, contributed as:
\[
\boldsymbol{\mu}_{\text{op\_in},i}^{(k)}
= \sum_{j} \alpha_{i\leftarrow j}^{(k)} \,
\boldsymbol{\mu}_{\ell_{j\rightarrow i}}^{(k)},
\qquad
\boldsymbol{\sigma}_{\text{op\_in},i}^{(k)}
= \sum_{j} \left|\alpha_{i\leftarrow j}^{(k)}\right| \,
\boldsymbol{\sigma}_{\ell_{j\rightarrow i}}^{(k)}.
\]
Outgoing messages use $\alpha_{j\leftarrow i}^{(k)}$ with the same formulation. All four aggregated vectors are concatenated:
\[
\boldsymbol{\mu}_{\text{concat},i}^{(k)} =
\big[
\boldsymbol{\mu}_{\text{in},i}^{(k)} \;\|\;
\boldsymbol{\mu}_{\text{out},i}^{(k)} \;\|\;
\boldsymbol{\mu}_{\text{op\_in},i}^{(k)} \;\|\;
\boldsymbol{\mu}_{\text{op\_out},i}^{(k)}
\big],
\]
\[
\boldsymbol{\sigma}_{\text{concat},i}^{(k)} =
\big[
\boldsymbol{\sigma}_{\text{in},i}^{(k)} \;\|\;
\boldsymbol{\sigma}_{\text{out},i}^{(k)} \;\|\;
\boldsymbol{\sigma}_{\text{op\_in},i}^{(k)} \;\|\;
\boldsymbol{\sigma}_{\text{op\_out},i}^{(k)}
\big].
\]
Following concatenation, we apply a learnable linear transformation. To ensure the non-negativity of variance estimates, the weight matrix magnitude is applied to the standard deviation channel:
\[
\boldsymbol{\mu}_{i}^{(k)'} = \mathbf{W}^{(k)} \, \boldsymbol{\mu}_{\text{concat},i}^{(k)} + \mathbf{b}^{(k)},\quad
\boldsymbol{\sigma}_{i}^{(k)'}
= \left|\mathbf{W}^{(k)}\right| \, \boldsymbol{\sigma}_{\text{concat},i}^{(k)}.
\]
The mean undergoes a Rectified Linear Unit (ReLU) transformation. To maintain probabilistic consistency, the uncertainty is gated by an indicator function that collapses the standard deviation to zero for non-activated regions, reflecting the deterministic nature of ReLU in negative domains:
\[
\boldsymbol{\mu}_i^{(k)}
= \text{ReLU}\left(\boldsymbol{\mu}_i^{(k)'}\right),
\qquad
\boldsymbol{\sigma}_i^{(k)}
= \boldsymbol{\sigma}_i^{(k)'}
\circ \mathbb{I}\left(\boldsymbol{\mu}_i^{(k)'} > 0\right),
\]
where $\mathbb{I}(\cdot)$ denotes the indicator function that returns 1 when the condition is satisfied and 0 otherwise. This formulation ensures that uncertainty propagates only through activated neurons, consistent with the piecewise linear structure of ReLU.

when the RAECA robustness is disabled, the SL reverts to a uniform mean aggregation strategy, in which it treats all node neighbors symmetrically, assuming equal reliability across the local neighborhood structure.

\subsubsection{Multi-layer Node Embedding Refinement}
\label{sec:multi_layer_aggregation}

The initial Gaussian node embeddings $(\boldsymbol{\mu}^{(0)}, \boldsymbol{\sigma}^{(0)})$ are refined through a stack of $L$ Gaussian convolution layers. Each layer $k$ aggregates information in two complementary ways. First, direct aggregation collects information from immediate one-hop neighbors using $(\boldsymbol{\mu}^{(k-1)}, \boldsymbol{\sigma}^{(k-1)})$. Second, because each layer operates on embeddings already enriched by preceding layers, stacking yields effective multi-hop aggregation that propagates trust signals and uncertainty over longer relational paths. Each layer also incorporates uncertainty-aware edge information $(\boldsymbol{\mu}_\ell^{(k)}, \boldsymbol{\sigma}_\ell^{(k)})$ from Section~\ref{sec:edge_label_gaussian}.

Formally, the update process for the $k$-th layer can be abstracted as a recursive convolution where the output of one layer serves as the input to the next:
\begin{equation}
\label{eq:layer_update_abstract}
\mathbf{h}_i^{(k)} = \mathrm{ConvLayer}^{(k)}\Big( \mathbf{h}_i^{(k-1)}, \big\{ (\mathbf{h}_j^{(k-1)}, \mathbf{e}_{j\to i}^{(k)}) \big\}_{j \in \mathcal{N}_i} \Big)
\end{equation}
where $\mathbf{h}_i^{(k)} = (\boldsymbol{\mu}_i^{(k)}, \boldsymbol{\sigma}_i^{(k)})$ denotes the Gaussian node embedding, and $e_{j\to i}^{(k)} = (\boldsymbol{\mu}_{\ell_{j\to i}}^{(k)}, \boldsymbol{\sigma}_{\ell_{j\to i}}^{(k)})$ is the Gaussian edge embedding at layer $k$. This recursive structure expands the receptive field with each step, allowing the final representation $(\boldsymbol{\mu}_i^{(L)}, \boldsymbol{\sigma}_i^{(L)})$ to summarize information from the $L$-hop neighborhood.

\vspace{-2mm}
\subsection{The Temporal Layer Framework}
\label{sec:temporal_framework}
The Temporal Layer models the dynamic evolution of trust by capturing dependencies across the sequence of node embeddings generated by the Structural Layer. Let $\mathbf{X} \in \mathbb{R}^{N \times T \times d'}$ denote the input tensor obtained by stacking the structural NEs $\boldsymbol{\mu}_i^{(L)}$ across $T$ discrete snapshots, where $N$ is the number of nodes and $d'$ is the embedding dimension. To address both the discrete nature of sampled snapshots and the continuous nature of underlying trust dynamics, this framework employs a hybrid architecture. It processes the input sequence through a discrete encoding phase using nonlinear Chebyshev-KAN layer and attention mechanisms, followed by a continuous-time residual correction phase governed by ODEs.

\subsubsection{Discrete Temporal Encoding via KAN-Attention}
The discrete encoding phase transforms the static sequence of structural embeddings into temporally aware representations. This process integrates adaptive positional information with expressive nonlinear feature transformations, replacing standard linear layers with polynomial-based mappings to capture complex trust behaviors.

\paragraph{HAGH Positional Encoding}
To provide the model with explicit temporal context on a very limited temporal timeslots, we employ HAGH positional encoding. This mechanism superimposes learned absolute positions with modulated Gaussian and Hourglass functions to highlight specific intervals and periodicities within the trust sequence. A unique positional identity vector $\mathbf{p}_{t}$ at time step $t$ is computed as:
$$
\resizebox{\columnwidth}{!}{$
\mathbf{p}_{t} = \mathbf{A}[t,:]
+ \exp\left(-\frac{(t-\mu)^2}{2\sigma^2}\right)\mathbf{W}_{\mathrm{Gau}}
+ \left(1 - \frac{2|t-c|}{T-1}\right)\mathbf{W}_{\mathrm{Hour}}
$}
$$
Here, $\mathbf{A} \in \mathbb{R}^{T \times d'}$ represents the learnable absolute embedding matrix. The parameters $\mu$ and $\sigma$ control the center and width of the Gaussian kernel, while $\mathbf{W}_{\mathrm{Gau}}$ and $\mathbf{W}_{\mathrm{Hour}}$ are learnable scaling vectors. The term $c = (T-1)/2$ denotes the midpoint of the temporal window. This encoding is added to the structural input, yielding the time-enriched embedding $\mathbf{z}_n^{(t)} = \mathbf{x}_n^{(t)} + \mathbf{p}_t$.

\paragraph{Nonlinear Feature Transformation via Chebyshev KAN}
Standard linear projections are limited to modeling proportional relationships. To capture nonlinear trust dynamics such as acceleration (convex growth) or saturation (concave leveling), we utilize KAN based on Chebyshev polynomials. We specifically adopt a degree-2 expansion ($K=2$), which enables the layer to natively approximate parabolic curvature. The recursive definition for the Chebyshev polynomials $T_k(v)$ up to degree 2 is:
\begin{equation}
T_0(v) = 1, \quad T_1(v) = v, \quad T_2(v) = 2v^2 - 1.
\end{equation}
The output feature $v_{\mathrm{out},o}$ for the $o$-th output dimension is computed as a weighted sum of these basis functions applied to the input feature $\tilde{v}_{\mathrm{in},i}$:
\begin{equation}
v_{\mathrm{out},o} = \sum_{i=1}^{d_{\mathrm{in}}} \sum_{k=0}^2 \Theta_{i,o,k} \cdot T_k(\tilde{v}_{\mathrm{in},i}),
\end{equation}
where $d_{\mathrm{in}}$ is the input dimension and $\Theta \in \mathbb{R}^{d_{\mathrm{in}} \times d_{\mathrm{out}} \times 3}$ represents the learnable coefficient tensor containing weights for the constant, linear, and quadratic terms. This formulation replaces the weight matrix of a standard linear layer, allowing the model to learn feature-wise nonlinearities efficiently.

\paragraph{KAN-Projected Multi-Head Attention}
The attention mechanism aggregates historical information by computing similarity scores between time steps. We employ a multi-head architecture to allow the model to jointly attend to information from different representation subspaces at different positions; for instance, one head may focus on recent short-term fluctuations while another captures long-term stability. Unlike traditional Transformers that use linear projections, we generate the query ($\mathbf{q}$), key ($\mathbf{k}$), and value ($\mathbf{v}$) vectors for every snapshot $\tau$ using independent KAN layers. In this framework, the query $\mathbf{q}_n^{(t,h)}$ represents the node's focus at the current time $t$, while keys $\mathbf{k}_n^{(s,h)}$ and values $\mathbf{v}_n^{(s,h)}$ represent the content retrieved from historical snapshots $s \le t$. For a node $n$ and head $h$, these projections are defined for any time step $\tau$ as:
$$
\scalebox{0.80}{$
\mathbf{q}_n^{(t,h)}=\mathrm{KAN}_Q^{(h)}(\mathbf{z}_n^{(t)}),\;
\mathbf{k}_n^{(t,h)}=\mathrm{KAN}_K^{(h)}(\mathbf{z}_n^{(t)}),\;
\mathbf{v}_n^{(t,h)}=\mathrm{KAN}_V^{(h)}(\mathbf{z}_n^{(t)})
$}
$$
The attention weights $\alpha_n^{(t,h)}(s)$ are derived from the scaled dot-product similarity between the current query at $t$ and past keys at $s$, normalized via a causal softmax. The head-specific context vector $\mathbf{u}_n^{(t,h)}$ is subsequently computed as the weighted sum of past values:
$$
a_n^{(t,h)}(s) = \frac{\langle \mathbf{q}_n^{(t,h)}, \mathbf{k}_n^{(s,h)} \rangle}{\sqrt{d_h}}, \quad
\mathbf{u}_n^{(t,h)} = \sum_{s=0}^t \alpha_n^{(t,h)}(s) \mathbf{v}_n^{(s,h)}.
$$
The outputs from all heads are concatenated and passed through a final feed-forward KAN layer ($\mathrm{KAN}_O$) with a residual connection to produce the discrete temporal embedding $\mathbf{h}_n^{(t)}$.
\vspace{-0.3em}
$$
\mathbf{h}_n^{(t)} = \mathrm{KAN}_O\big(\mathrm{Concat}(\mathbf{u}_n^{(t,1)},\dots,\mathbf{u}_n^{(t,H)})\big) + \mathbf{z}_n^{(t)}.
$$

\subsubsection{Continuous Residual Dynamics via Neural ODEs}
\label{sec:ode_dynamics}
The discrete attention mechanism effectively captures dependencies between observed snapshots but may miss hidden dynamics occurring between irregular sampling intervals. To address this, we introduce a residual pathway modeled by Neural ODEs.

\paragraph{Latent Residual Definition}
We define the residual $r_i(t_k)$ as the discrepancy between the spatially-grounded structural embedding $z_i^{\text{struct}}(t_k)$ (from the Structural Layer) and the history-based temporal prediction $z_i^{\text{temp}}(t_k)$ (from the KAN-Attention block) at snapshot $t_k$:
\begin{equation}
r_i(t_k) = z_i^{\text{struct}}(t_k) - z_i^{\text{temp}}(t_k).
\label{eq:residual_def}
\end{equation}
This residual encapsulates the *spatial innovations*—new trust information present in the graph snapshot that could not be inferred solely from past history. We treat this sequence $\{r_i(t_k)\}$ as observations of an underlying continuous latent process $h_i(t)$ governed by a data-conditioned ODE:
\begin{equation}
\frac{d h_i(t)}{d t} = g_\phi\big(h_i(t), r_i^{\text{obs}}(t)\big), \quad h_i(t_1) = r_i(t_1),
\label{eq:node_ode}
\end{equation}
where $g_\phi$ is a neural network parameterizing the derivative and $r^{\text{obs}}_i(t)$ is the piecewise-constant interpolation of the discrete residuals $\{r_i(t_k)\}_{k=1}^{N_T}$ that are used as exogenous input to the ODE. $N_T$ is the number of residual time points (snapshots) used in ODE fitting.

\paragraph{Integration and Refinement}
The continuous trajectory is obtained by numerically integrating Equation \ref{eq:node_ode} over the snapshot interval $[t_1, t_{N_T}]$ using an explicit Adams-Bashforth solver. This yields a smoothed latent trajectory $\hat{h}_i(t)$, which serves as a correction term. The final refined node embedding $\mathbf{Z}_i(t_k)$ is obtained by injecting this continuous residual back into the temporal representation:
\begin{equation}
\mathbf{Z}_i(t_k) = z_i^{\text{temp}}(t_k) + \hat{h}_i(t_k).
\end{equation}
This formulation compels the ODE module to learn the continuous dynamics of the prediction error, effectively bridging the gap between historical expectation and spatial reality.

\paragraph{Training Objective}
The model is optimized end-to-end using a composite loss function that balances task performance with residual consistency. The total loss $\mathcal{L}$ is defined as:
\begin{equation}
\mathcal{L} = \mathcal{L}_{\text{task}} + \lambda_{\text{res}} \cdot \mathcal{L}_{\text{res}},
\label{eq:total_loss}
\end{equation}
where $\mathcal{L}_{\text{task}}$ represents the primary supervised objective (e.g., cross-entropy for trust link prediction). The residual loss term $\mathcal{L}_{\text{res}}$ ensures that the generated continuous trajectory $\hat{h}_i(t)$ faithfully approximates the observed discrete residuals $r_i(t_k)$:
\begin{equation}
\mathcal{L}_{\text{res}} = \frac{1}{|V| N_T} \sum_{i \in V} \sum_{k=1}^{N_T} \big\| \hat{h}_i(t_k) - r_i(t_k) \big\|_2^2.
\label{eq:residual_loss}
\end{equation}
The hyperparameter $\lambda_{\text{res}} = 0.3$ controls the weighting of the residual loss relative to the main task. This auxiliary objective ensures that the ODE captures meaningful temporal patterns in the residuals rather than drifting arbitrarily, thereby stabilizing the training of the continuous dynamics.

\paragraph{Integration and Refinement}
The continuous trajectory is obtained by numerically integrating Equation \ref{eq:node_ode} over the snapshot interval $[t_1, t_{N_T}]$ using an explicit Adams-Bashforth solver. This yields a smoothed latent trajectory $\hat{h}_i(t)$, which serves as a correction term. The final refined node embedding $\mathbf{Z}_i(t_k)$ is obtained by injecting this continuous residual back into the temporal representation:
\begin{equation}
\mathbf{Z}_i(t_k) = z_i^{\text{temp}}(t_k) + \hat{h}_i(t_k).
\end{equation}
This formulation compels the ODE module to learn the hidden continuous dynamics that bridge the discrete temporal predictions, providing a refined representation that respects both historical trends and instantaneous structural constraints.

\vspace{-2mm}
\subsection{Prediction Layer and Optimization}
The final refined node embeddings, represented by the tensor $\mathbf{Z}$ (where $\mathbf{Z}[n, t, :]$ denotes the embedding for node $n$ at time step $t$), serve as the foundation for downstream trust/distrust predictions. To infer the relationship between two nodes $n$ and $p$ at a specific time step $t_k$, we extract their corresponding time-specific embeddings $\mathbf{Z}[n, t_k, :]$ and $\mathbf{Z}[p, t_k, :]$. These embeddings are fed into a prediction head, which combines them through concatenation followed by a linear transformation to produce a vector of logits $\hat{\mathbf{y}}_{np, t_k} \in \mathbb{R}^C$, where $C$ represents the number/level of trust classes.

DGTEN is trained end-to-end by minimizing a weighted Cross-Entropy loss. Let $\mathcal{D}_{\text{train}}$ be the set of training instances, and let $y_{np, t_k} \in \{0, 1\}$ be the ground-truth class label. To address class imbalance, we scale the loss for each instance using a weight $\alpha_{y}$ derived from the class frequencies:
$ \alpha_{y} = \frac{|\mathcal{D}_{\mathrm{train}}|}{\text{count}(y)}, $
where $\text{count}(y)$ is the number of training samples belonging to class $y$. The task objective is defined as:
\begin{equation}  
\label{eq:loss_function_cross_entropy}
\mathcal{L}_{\text{task}} = - \sum_{(n,p, y, t_k) \in \mathcal{D}_{\text{train}}} \alpha_{y} \log \left( \frac{\exp(\hat{\mathbf{y}}_{np, t_k}[y])}{\sum_{j=0}^{1} \exp(\hat{\mathbf{y}}_{np, t_k}[j])} \right)
\end{equation}
The model parameters $\Phi$ are optimized to minimize $\mathcal{L}_{\text{}}$ using the MadGrad optimizer \cite{defazio2022momentumized}. We employ weight decay (set to $10^{-5}$) on the model parameters to prevent overfitting, which is integrated directly into the optimizer's update step.
\vspace{-2mm}
\subsection{Computational Complexity Analysis}
We analyze the forward-pass time complexity($\mathcal{O}$) of DGTEN, where $N = |V|$ denotes nodes, $T$ temporal snapshots, $L$ DGMP layers, $|E|$ edges per snapshot, $d$ embedding dimension, and $H$ attention heads. Each of the $L$ Structural Layer convolutions processes $|E|$ edges across $T$ snapshots, where RAECA similarity computation requires $\mathcal{O}(|E| d)$ operations and Gaussian aggregation with linear projections costs $\mathcal{O}(|E| d + V d^2)$ per layer. For sparse graphs where $|E| = \mathcal{O}(V)$, the total Structural Layer complexity is $\mathcal{O}(T L |E| d)$.

The Temporal Layer performs KAN-based projections costing $\mathcal{O}(T d^2)$ per node, followed by multi-head causal self-attention that computes pairwise temporal similarities requiring $\mathcal{O}(N T^2 d)$ operations. Neural ODE integration using the Adams-Bashforth solver evaluates the derivative network across $T$ intervals, contributing an additional $\mathcal{O}(N T d^2)$ cost. Combining both layers, the end-to-end forward-pass complexity is
\begin{equation}
\mathcal{O}(T L |E| d + N T^2 d + N T d^2),
\end{equation}


\section{Experimental Design and Results}
\label{sec:experiments}
\subsection{Experimental Setup}
\begin{figure*}[!htbp]
    \centering
    \includegraphics[width=0.38\textwidth]{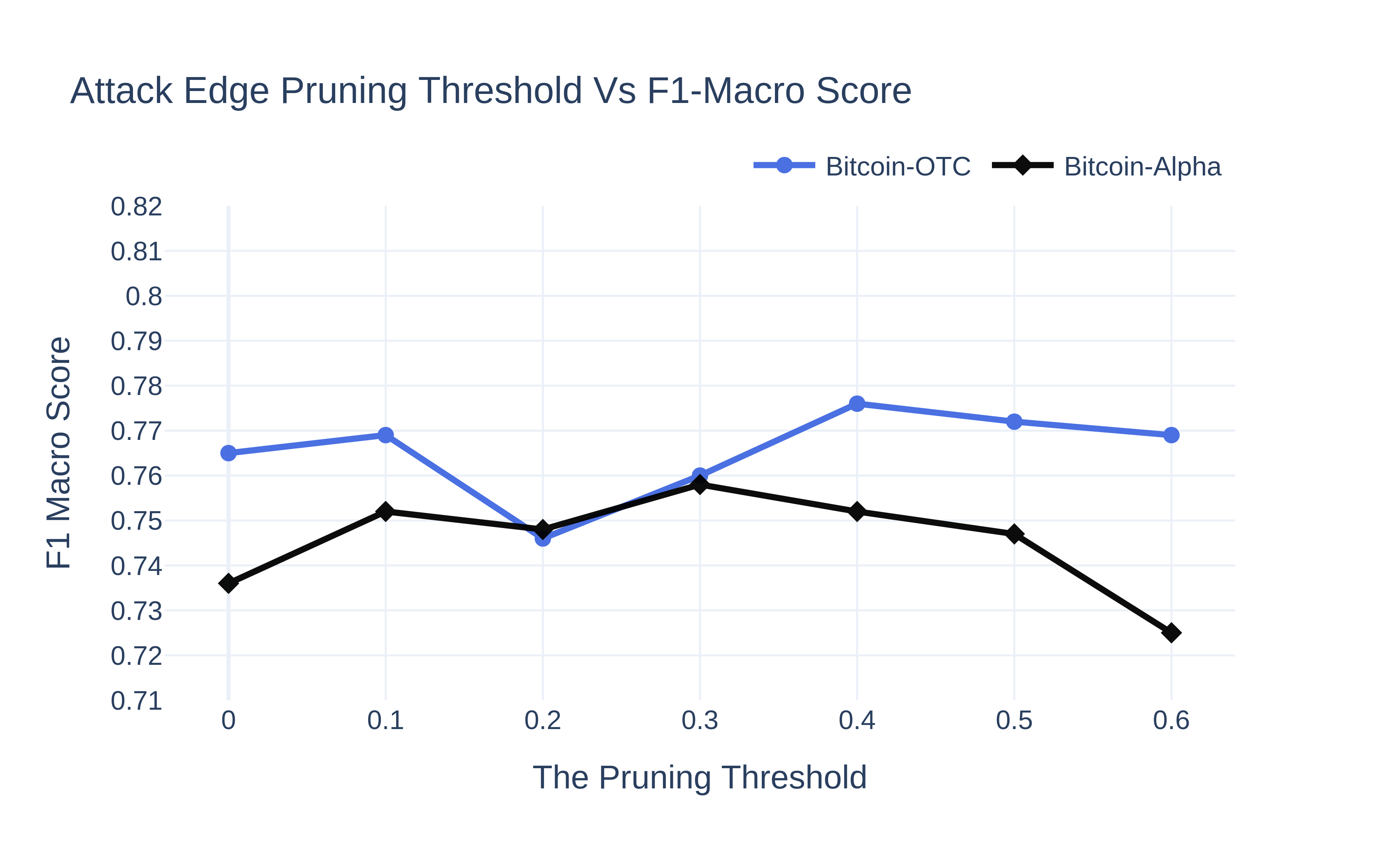}
    \hfill
    \includegraphics[width=0.58\textwidth]{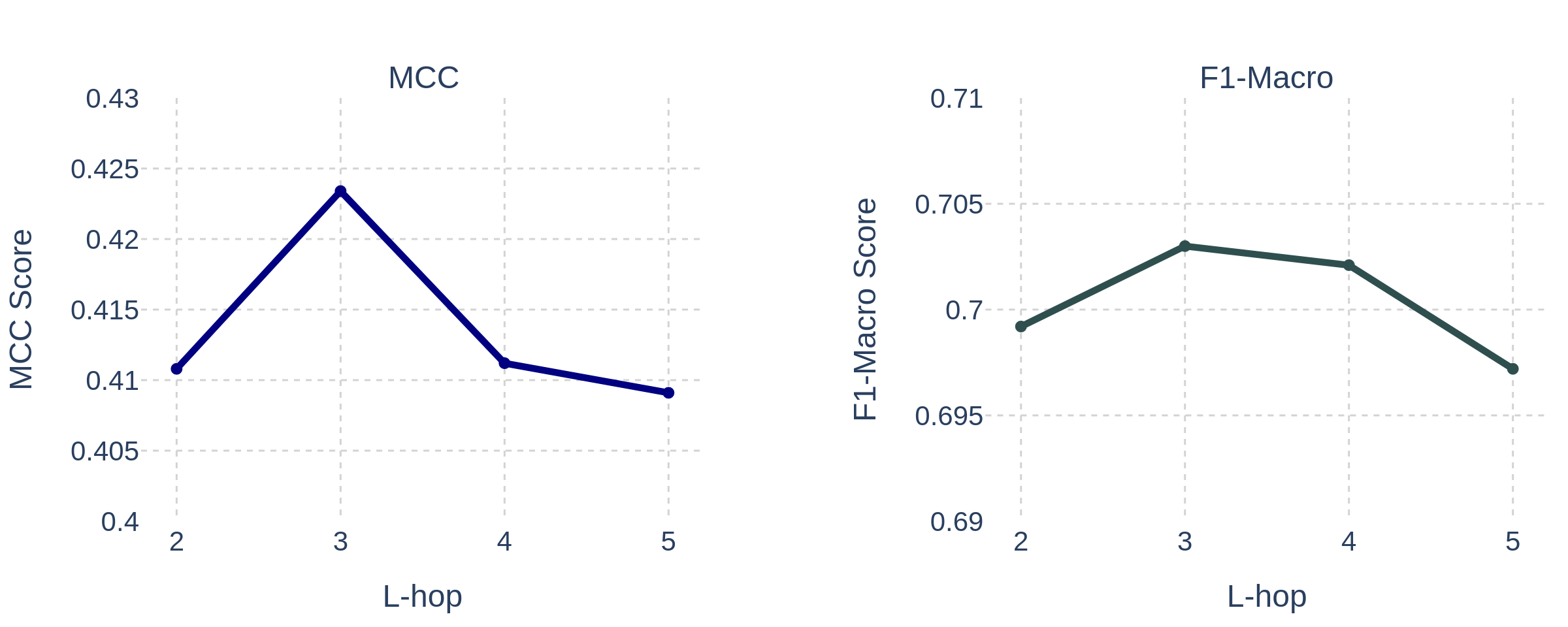}
    \vspace{-0.33mm}
    \caption{Hyperparameter analysis for DGTEN. Left: F1-Macro vs. Pruning Threshold ($\tau$). Right: Effect of L-hop Depth.}
    \label{fig:hyperparams}
    \vspace{-0.8em}
\end{figure*}

DGTEN is implemented in PyTorch and trained on a workstation equipped with 32 CPU cores and an NVIDIA A6000 GPU. Hyperparameters are selected through grid search, while all remaining settings follow established configurations from prior studies. The search space includes the L hop convolution depth (Fig.~\ref{fig:hyperparams}), the RAECA pruning threshold (Fig.~\ref{fig:hyperparams}), the learning rate (lr), weight decay (wd), and the temporal attention configuration. Unless otherwise stated, training is conducted with $\text{lr}=0.005$, $\text{wd}=10^{-5}$, and $250$ epochs. A convolution depth of $\text{L}=3$, consistent with prior work~\cite{10.5555/3294771.3294869, huo2023trustgnn}, yields the most reliable overall performance.

Grid search further identifies dataset specific cosine similarity pruning thresholds of $\tau=1.4$ for BitcoinOTC and $\tau=1.3$(Fig.~\ref{fig:hyperparams}) for BitcoinAlpha. The temporal attention module employs an attention dropout rate of 0.5, with 16 heads for Bitcoin OTC and 8 heads for Bitcoin Alpha. Attack configurations and robust aggregation settings vary across experiments and are reported together with the corresponding results.

For static dataset experiments, we again perform an analogous grid search to determine dataset specific hyperparameters. For the PGP dataset with a 40\% training split, we found $\text{lr}=0.01$, $\text{wd}=10^{-6}$, a dropout rate of $0.3$, and $300$ training epochs; increasing the training proportion by 20\% results in a proportional increase in the number of epochs. For Advogato with a 40\% training split, we use $\text{lr}=0.003$, $\text{wd}=10^{-5}$, no dropout, and $200$ epochs, with epoch counts scaled proportionally when the training set size increases.

\subsection{DGTEN Evaluation Protocol}
We evaluate DGTEN using an expanding-window protocol over the dynamic graph $\mathcal{G} = \{G_1, G_2, \ldots, G_N\}$, where $N$ is the total number of snapshots and snapshots are ordered chronologically.

The initial training window $T_{\text{initial}}$ contains the number of initial snapshots used to form the first training window (e.g., $T_{\text{initial}} = 2$). As at least 2 snapshots are required for our temporal modeling. In each evaluation round, the model trains on $
\mathcal{G}_{\text{train}} = \{G_1, G_2, \ldots, G_{t_{\text{end}}}\}, $
where $t_{\text{end}}$ starts at $T_{\text{initial}}$ and increases by one each round. 
Each round constitutes an independent experiment with fresh model initialization and fixed training epochs. During training, we save the model parameters that achieve the lowest training loss and load these best parameters for evaluation. The model then predicts trust-activities (edges) in future snapshots after $G_{t_{\text{end}}}$. For single-timeslot 
prediction, this yields $N - T_{\text{initial}}$ rounds, from predicting 
$G_{T_{\text{initial}}+1}$ through $G_N$.
Averaging metrics across all rounds provides the overall experimental result and a stable measure of generalization as temporal information grows. This evaluation procedure applies to three tasks:

\vspace{0.5\baselineskip}
\noindent \textit{Task 1}: Single-Timeslot Prediction (Observed Nodes).
In each round, the model trains on the sequence
$\{G_1, \ldots, G_{t_{\text{end}}}\}$, where
$t_{\text{end}} \in [T_{\text{initial}},\, N-1]$,
and predicts the next snapshot $G_{t_{\text{end}}+1}$.
The prediction set includes observed nodes, i.e., all entities
that appear in $\{G_1, \ldots, G_{t_{\text{end}}}\}$.

\vspace{0.5\baselineskip}

\noindent \textit{Task 2}: Multi-Timeslot Prediction (Observed Nodes).
After training on $\{G_1, \ldots, G_{t_{\text{end}}}\}$, the model predicts the sequence
$\{G_{t_{\text{end}}+1}, G_{t_{\text{end}}+2}, \ldots, G_{t_{\text{end}}+\Delta}\}$,
where $\Delta > 1$ (we use $\Delta = 3$) and
$t_{\text{end}} \in [T_{\text{initial}},\, N-\Delta]$.
Prediction targets remain the observed nodes.

\vspace{0.5\baselineskip}

\noindent \textit{Task 3}: Single-Timeslot Prediction (Unobserved Nodes).
In this cold-start setting, the model trains on
$\{G_1, \ldots, G_{t_{\text{end}}}\}$, where
$t_{\text{end}} \in [T_{\text{initial}},\, N-1]$,
and predicts $G_{t_{\text{end}}+1}$.
Targets are unobserved nodes, i.e., entities absent from
$\{G_1, \ldots, G_{t_{\text{end}}-1}\}$ that appear for the first time in $G_{t_{\text{end}}}$.

\vspace{0.5\baselineskip}

We run each experiment five times independently for every task (Task-1, Task-2, and Task-3) to obtain statistically stable estimates. All reported metrics, including those measured under adversarial attack settings, are computed as the mean±std over these five runs. This protocol enables compact yet comprehensive evaluation of immediate prediction, multi-step forecasting, and cold-start performance. Results compared with state-of-the-art 
baselines appear in Table~\ref{table:comparison}.
\vspace{-2mm}
\subsection{Performance Metrics}
We evaluate DGTEN in imbalanced trust graphs using six metrics. The terms {TP}, {TN}, {FP}, and {FN} denote true/false positives/negatives, where \textit{distrust} is the positive class ($P$) and \textit{}{trust} is the negative class ($N$). All metrics, except MCC (range $-1$ to $+1$), span $0$ to $1$.

\subsubsection*{1. Area Under the ROC Curve (AUC)}
Quantifies the model's overall {discriminative ability} across all thresholds. It is computed from the ranks of positive instances:
\[
\displaystyle
\mathrm{AUC} = 
\frac{\sum_{i \in P} \mathrm{rank}_i - |P|(|P|+1)/2}{|P|\cdot|N|}
\]
where $|P|$ and $|N|$ are the counts of positive and negative instances, and $\mathrm{rank}_i$ is the rank of the $i$-th positive instance based on predicted scores.

\subsubsection*{2. Matthews Correlation Coefficient (MCC)}
A {reliable measure for imbalanced datasets}, providing a balanced score for binary classification quality:
\[
\displaystyle
\mathrm{MCC} = 
\frac{TP \times TN - FP \times FN}
{\sqrt{(TP + FP)(TP + FN)(TN + FP)(TN + FN)}}
\]

\subsubsection*{3. Balanced Accuracy (BA)}
Addresses imbalance by averaging {sensitivity} (True Positive Rate) and {specificity} (True Negative Rate):
\[
\displaystyle
\mathrm{BA} = 
\frac{1}{2}\left(\frac{TP}{TP + FN} + \frac{TN}{TN + FP}\right)
\]

\subsubsection*{4. Average Precision (AP)}
Summarizes the {precision-recall curve} through a weighted mean of precision at varying recall levels:
\[
\displaystyle
\mathrm{AP} = \sum_{k}(R_k - R_{k-1})P_k
\]
where $P_k$ and $R_k$ are precision and recall at the $k$-th threshold.
\subsubsection*{5. Micro-Averaged F1 Score (F1-Micro)}
Aggregates \(TP\), \(FP\), \(FN\) globally across classes:
\[
\displaystyle
\mathrm{F1\text{-}Micro} = 
2 \cdot 
\frac{
\frac{\sum TP}{\sum TP + \sum FP} \cdot 
\frac{\sum TP}{\sum TP + \sum FN}
}{
\frac{\sum TP}{\sum TP + \sum FP} + 
\frac{\sum TP}{\sum TP + \sum FN}
}
\]
\subsubsection*{6. Macro-Averaged F1 Score (F1-Macro)}
Calculates the F1 score for each class ($c \in C$) independently and then averages them (unweighted), giving {equal importance to each class} regardless of size:
\[
\displaystyle
\mathrm{F1\text{-}Macro} = 
\frac{1}{|C|}
\sum_{c \in C}\mathrm{F1}_c
\]

\begin{table*}[!t]
\centering
\caption{DGTEN performance(mean $\pm$ std) on clean data with RAECA enabled. Best results in bold}
\label{table:comparison}
\setlength{\tabcolsep}{3.5pt} 
\renewcommand{\arraystretch}{0.95} 
\resizebox{\textwidth}{!}{%
\begin{tabular}{llcccccccc}
\toprule
\multirow{2}{*}{\textbf{Task}} & \multirow{2}{*}{\textbf{Model}} & \multicolumn{4}{c}{\textbf{Bitcoin-OTC}} & \multicolumn{4}{c}{\textbf{Bitcoin-Alpha}} \\
\cmidrule(lr){3-6} \cmidrule(lr){7-10}
 & & \textbf{MCC} & \textbf{AUC} & \textbf{BA} & \textbf{F1-Macro} & \textbf{MCC} & \textbf{AUC} & \textbf{BA} & \textbf{F1-Macro} \\
\midrule
\multirow{4}{*}{Task-1} 
 & Guardian \cite{li2020guardian} & $0.351\pm0.007$ & $0.744\pm0.003$ & $0.653\pm0.006$ & $0.668\pm0.002$ & $0.328\pm0.012$ & $0.734\pm0.009$ & $0.671\pm0.004$ & $0.649\pm0.004$ \\
 & GATrust \cite{gao2020gatrust} & $0.346\pm0.005$ & $0.743\pm0.004$ & $0.654\pm0.008$ & $0.660\pm0.007$ & $0.329\pm0.009$ & $0.729\pm0.004$ & $0.663\pm0.004$ & $0.648\pm0.010$ \\
 & TrustGuard \cite{wang2024trustguard} & $0.389\pm0.007$ & $0.765\pm0.008$ & $0.693\pm0.008$ & $0.687\pm0.007$ & $0.362\pm0.004$ & $0.756\pm0.009$ & $0.692\pm0.004$ & $0.669\pm0.002$ \\
 
 & \textbf{DGTEN(ours)} & 
 
 $\mathbf{0.437\pm0.012}$ & $\mathbf{0.795\pm0.003}$ & $\mathbf{0.702\pm0.004}$ & $\mathbf{0.715\pm0.004}$ & 
 $\mathbf{0.405\pm0.004}$ & $\mathbf{0.785\pm0.003}$ & $\mathbf{0.717\pm0.010}$ & $\mathbf{0.690\pm0.005}$ \\
 
\cmidrule{2-10}
 & \textbf{Improv.(\%)} & +12.34\% & +3.92\% & +1.30\% & +4.08\% & +11.88\% & +3.84\% & +3.61\% & +3.14\% \\
\midrule
\multirow{4}{*}{Task-2} 
 & Guardian \cite{li2020guardian} & $0.295\pm0.005$ & $0.715\pm0.003$ & $0.622\pm0.001$ & $0.639\pm0.001$ & $0.260\pm0.006$ & $0.673\pm0.003$ & $0.618\pm0.005$ & $0.624\pm0.003$ \\
 & GATrust \cite{gao2020gatrust} & $0.290\pm0.002$ & $0.714\pm0.002$ & $0.622\pm0.002$ & $0.636\pm0.003$ & $0.257\pm0.004$ & $0.675\pm0.003$ & $0.615\pm0.005$ & $0.621\pm0.002$ \\
 & TrustGuard \cite{wang2024trustguard} & $0.330\pm0.005$ & $0.725\pm0.004$ & $0.642\pm0.003$ & $0.658\pm0.003$ & $0.288\pm0.002$ & $0.692\pm0.003$ & $0.632\pm0.006$ & $0.639\pm0.001$ \\
 
 & \textbf{DGTEN(ours)} & 
 $\mathbf{0.362\pm0.003}$ & $\mathbf{0.750\pm0.003}$ & $\mathbf{0.649\pm0.006}$ & $\mathbf{0.671\pm0.002}$ & $\mathbf{0.317\pm0.004}$ & $\mathbf{0.723\pm0.005}$ & $\mathbf{0.648\pm0.006}$ & $\mathbf{0.646\pm0.003}$ \\
\cmidrule{2-10}
 & \textbf{Improv.(\%)} & +9.70\% & +3.45\% & +1.09\% & +1.98\% & +10.07\% & +4.48\% & +2.53\% & +1.10\% \\
\midrule
\multirow{4}{*}{Task-3} 
 & Guardian \cite{li2020guardian} & $0.447\pm0.019$ & $0.709\pm0.016$ & $0.667\pm0.004$ & $0.693\pm0.005$ & $0.325\pm0.012$ & $0.678\pm0.015$ & $0.631\pm0.010$ & $0.641\pm0.005$ \\
 & GATrust \cite{gao2020gatrust} & $0.430\pm0.014$ & $0.712\pm0.011$ & $0.672\pm0.006$ & $0.691\pm0.006$ & $0.321\pm0.008$ & $0.681\pm0.014$ & $0.627\pm0.008$ & $0.636\pm0.004$ \\
 & TrustGuard \cite{wang2024trustguard} & $0.463\pm0.020$ & $0.727\pm0.014$ & $0.673\pm0.009$ & $0.701\pm0.009$ & $0.384\pm0.026$ & $0.715\pm0.027$ & $0.654\pm0.012$ & $0.678\pm0.013$ \\
 & \textbf{DGTEN(ours)} & 
 
 $\mathbf{0.483\pm0.005}$ & $\mathbf{0.734\pm0.012}$ & $\mathbf{0.685\pm0.003}$ & $\mathbf{0.714\pm0.004}$ & $\mathbf{0.480\pm0.008}$ & $\mathbf{0.725\pm0.006}$ & $\mathbf{0.711\pm0.009}$ & $\mathbf{0.725\pm0.004}$ \\
\cmidrule{2-10}
& \textbf{Improv.(\%)} & +4.32\% & +0.96\% & +1.78\% & +1.85\% & +25.00\% & +1.40\% & +8.72\% & +6.93\% \\
\bottomrule
\end{tabular}%
}
\vspace{-3mm}
\end{table*}

\begin{table}[t]
\centering
\caption{DGTEN's SL performance on static datasets with varying training percentages (40\%, 60\%, 80\%). Best results in bold.}
\label{tab:performance_comparison}
\resizebox{\columnwidth}{!}{%
\begin{tabular}{@{}l *{6}{c}@{}}
\toprule
\multirow{2}{*}{Methods} 
& \multicolumn{3}{c}{Advogato (Micro-F1)} 
& \multicolumn{3}{c}{PGP (Micro-F1)} \\
\cmidrule(lr){2-4} \cmidrule(lr){5-7}
& 40\% & 60\% & 80\% & 40\% & 60\% & 80\% \\
\midrule
MoleTrust~\cite{massa2005controversial}   & ---    & ---    & 58.4\% & ---    & ---    & 64.0\% \\
AssessTrust~\cite{liu2014assessment}     & ---    & ---    & 63.9\% & ---    & ---    & ---    \\
Matri~\cite{yao2013matri}                 & 61.7\% & 63.9\% & 65.0\% & 60.5\% & 64.7\% & 67.3\% \\
OpinionWalk~\cite{liu2017opinionwalk}     & ---    & ---    & 63.3\% & ---    & ---    & 66.8\% \\
NeuralWalk~\cite{liu2019neuralwalk}       & ---    & ---    & 74.0\% & ---    & ---    & ---    \\
Guardian                                   & 69.7\% & 71.7\% & 73.0\% & 84.6\% & 85.9\% & 86.7\% \\
TrustGNN~\cite{huo2023trustgnn}           & 70.1\% & \textbf{72.6\%} & \textbf{74.6\%} & \textbf{85.4\%} & 86.3\% & 87.2\% \\
\midrule
\textbf{DGTEN}                             & \textbf{70.15\%} & 71.76\% & 73.25 & 85.32\% & \textbf{86.54\%} & \textbf{87.23\%} \\
\bottomrule
\end{tabular}%
}
\vspace{-0.3em}
\end{table}
\vspace{-8mm}
\subsection{Datasets}
To comprehensively evaluate our model, we utilized two dynamic signed networks, {Bitcoin-OTC} and {Bitcoin-Alpha}~\cite{7837846}, alongside two static trust datasets: Advogato\cite{massa2009bowling} and Pretty-Good-Privacy(PGP)\cite{rossi2015network}. The Bitcoin datasets capture temporal "who-trusts-whom" interactions from trading platforms, featuring weighted, signed edges with associated timestamps. For static comparisons, we include Advogato and PGP, which characterize trust into four hierarchical levels (Observer, Apprentice, Journeyer, Master). Table~\ref{tab:dataset_statistics} provides the summary statistics for all datasets.
\begin{table}[t]
\centering
\caption{Statistics of Datasets.}
\label{tab:dataset_statistics}
\resizebox{\columnwidth}{!}{%
\begin{tabular}{@{}lccccc@{}}
\toprule
\textbf{Dataset} & \textbf{Nodes} & \textbf{Edges} & \textbf{Pos. Ratio} & \textbf{Domain} & \textbf{Time Span} \\ 
\midrule
\multicolumn{6}{l}{\textit{Dynamic Networks (Weighted, Signed, Directed, Sparse)}} \\ 
Bitcoin-OTC & 5,881 & 35,592 & $\sim$90.0\% & Cryptocurrency & 2010--2016 \\
Bitcoin-Alpha & 3,775 & 24,186 &  $\sim$93.7\% & Cryptocurrency & 2010--2016 \\ 
\midrule
\multicolumn{6}{l}{\textit{Static Networks (Levels of Trust/Distrust Relations)}} \\ 
Advogato & 6,541 & 51,127 & --- & Software Dev. & Static \\
PGP & 38,546 & 317,979 & --- & Key Certification  & Static \\
\bottomrule

\end{tabular}%
}
\vspace{-1.0em}
\end{table}

\subsection{Baseline Models}
To provide a thorough evaluation, we incorporate both dynamic and static baselines. TrustGuard~\cite{wang2024trustguard}, representing the current state-of-the-art dynamic model, allows for a direct comparison on our temporal datasets. For reference, we also include leading static models, Guardian~\cite{li2020guardian} and GATrust~\cite{gao2020gatrust}, which overlook temporal dynamics. DGTEN's consistent outperformance of both model types highlights the advantages of incorporating dynamic trust elements and reveals the limitations of methods that disregard time-based factors. To further contextualize our model's performance, DGTEN is also evaluated against a suite of influential static baselines from the trust evaluation literature: TrustGNN\cite{huo2023trustgnn}, NeuralWalk\cite{liu2019neuralwalk}, OpenWalk\cite{liu2017opinionwalk}, Matri\cite{yao2013matri}, AssessTrust \cite{liu2014assessment}, and MoleTrust~\cite{massa2005controversial}.

\subsection{Comparative Performance Analysis}
To evaluate DGTEN's dynamic trust prediction capability, experiments were conducted on the Bitcoin-OTC and Bitcoin-Alpha datasets across three tasks. Table~\ref{table:comparison} reports the detailed results for DGTEN and the baselines Guardian, GATrust, and TrustGuard; the discussion emphasizes relative trends and percentage gains, highlighting DGTEN's capabilities in modeling temporal trust dynamics.

Task 1 evaluates prediction performance on observed nodes within a single timeslot(snapshot). On both datasets, DGTEN consistently outperforms all baselines across MCC, AUC, balanced accuracy, and F1-macro. On Bitcoin-OTC, DGTEN improves MCC by 12.34\% over TrustGuard, along with gains of 3.92\% in AUC and 4.08\% in F1-macro; on Bitcoin-Alpha, it yields an 11.88\% MCC increase, with AUC and F1-macro improvements of 3.84\% and 3.14\%, respectively, indicating strong short-term trust prediction performance in both networks. Task 2 assesses each model's ability to forecast trust over multiple future timeslots, a more challenging temporal setting. On Bitcoin-OTC, DGTEN improves MCC by 9.70\% over TrustGuard, with additional gains of 3.45\% (AUC), 1.09\% (balanced accuracy), and 1.98\% (F1-macro), while Guardian and GATrust remain consistently weaker. On Bitcoin-Alpha, DGTEN achieves even larger relative improvements, with a 10.07\% increase in MCC and gains of 4.48\% in AUC, 2.53\% in balanced accuracy, and 1.10\% in F1-macro over TrustGuard, underscoring the effectiveness of its temporal modeling components.

Task 3 focuses on cold-start prediction for previously unseen nodes, where models must infer trust without node-specific history. On Bitcoin-OTC, DGTEN achieves a 4.32\% MCC improvement over TrustGuard, while also increasing AUC by 0.96\%, balanced accuracy by 1.78\%, and F1-macro by 1.85\%, indicating more reliable predictions under limited information. On Bitcoin-Alpha, DGTEN exhibits its largest relative gains, with a 25.00\% MCC increase and improvements of 1.40\% (AUC), 8.72\% (balanced accuracy), and 6.93\% (F1-macro) over TrustGuard, highlighting strong generalization in the presence of epistemic uncertainty on unseen nodes. Cross-dataset comparison reveals that Bitcoin-Alpha tends to yield larger relative improvements for DGTEN, most notably in cold-start prediction, where the MCC gain reaches 25.00\% compared to 4.32\% on Bitcoin-OTC. Single-timeslot and multi-timeslot MCC improvements are also consistently high on both datasets (roughly 12.34\% vs.\ 11.88\% for Task 1 and 9.70\% vs.\ 10.07\% for Task 2 on Bitcoin-OTC and Bitcoin-Alpha, respectively), suggesting that DGTEN adapts well to different dynamic trust environments.

\subsubsection{Analysis of Static Performance}
To assess the structural node embedding capability of DGTEN, we evaluate the SL component in isolation on the Advogato and PGP datasets. Since these datasets are static graphs, temporal attention mechanisms and ODE-based components are omitted. This ablation isolates the effectiveness of the DGMP-based embedding module in capturing structural trust relationships without temporal information.

The results in Table~\ref{tab:dataset_statistics} show that the SL component delivers performance competitive with state-of-the-art baselines. On Advogato, DGTEN attains the highest Micro-F1 score at the 40\% training split (70.15\%), indicating strong embedding quality under limited supervision, although it performs slightly below TrustGNN at larger training ratios. In contrast, on the PGP dataset, DGTEN’s SL module outperforms all baselines at the 60\% and 80\% training splits, achieving 86.54\% and 87.23\%, respectively. These results suggest that the Gaussian message passing architecture effectively encodes trust semantics from graph topology, demonstrating strong structural learning capability even without temporal modeling. The multiclass cross entropy objective $\mathcal{L}_{\text{task}}$ in Eq.~\ref{eq:loss_function_cross_entropy} is used for training.

\begin{figure*}[!htbp]
    \centering
    \includegraphics[width=0.49\textwidth]{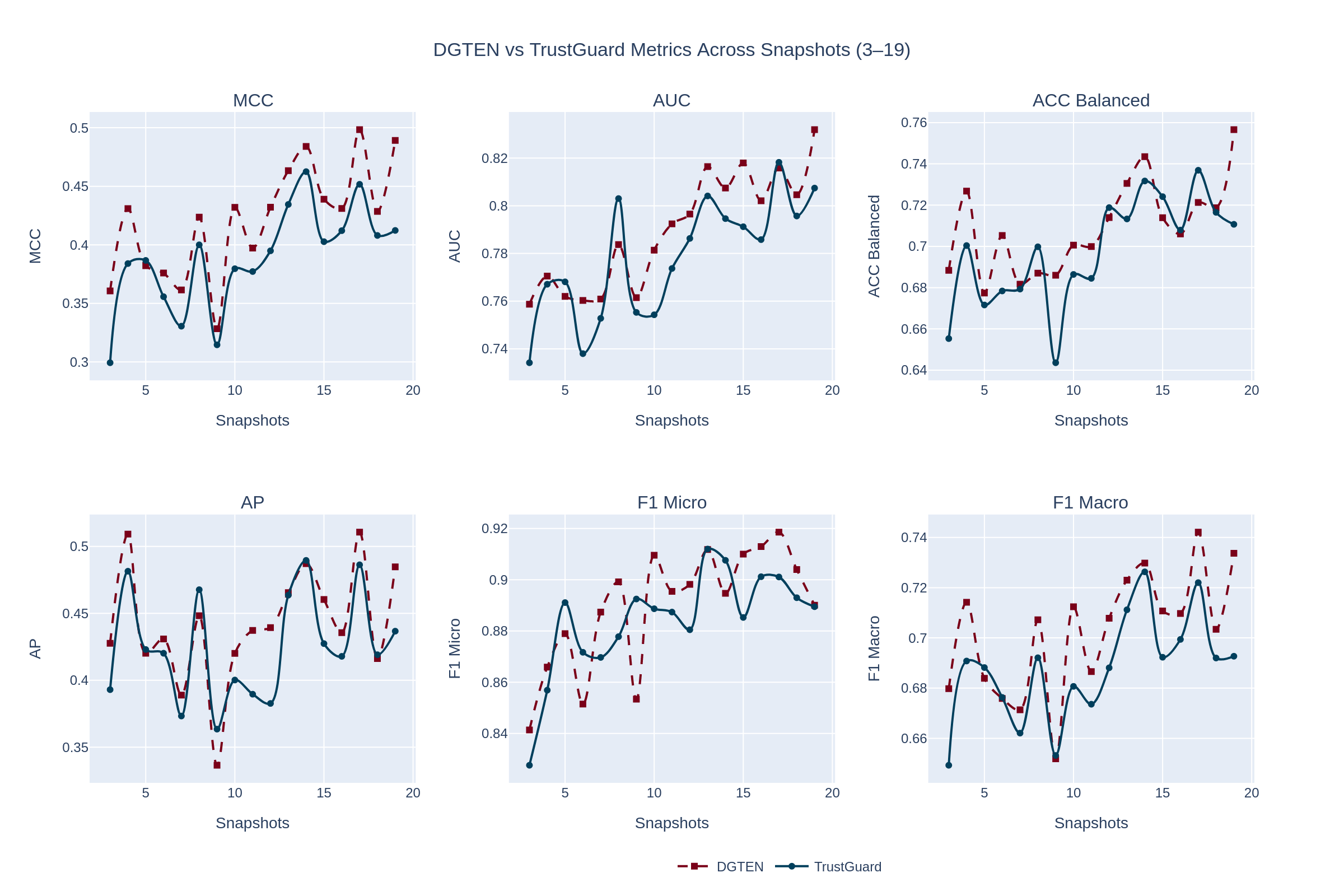}
    \hfill
    \includegraphics[width=0.49\textwidth]{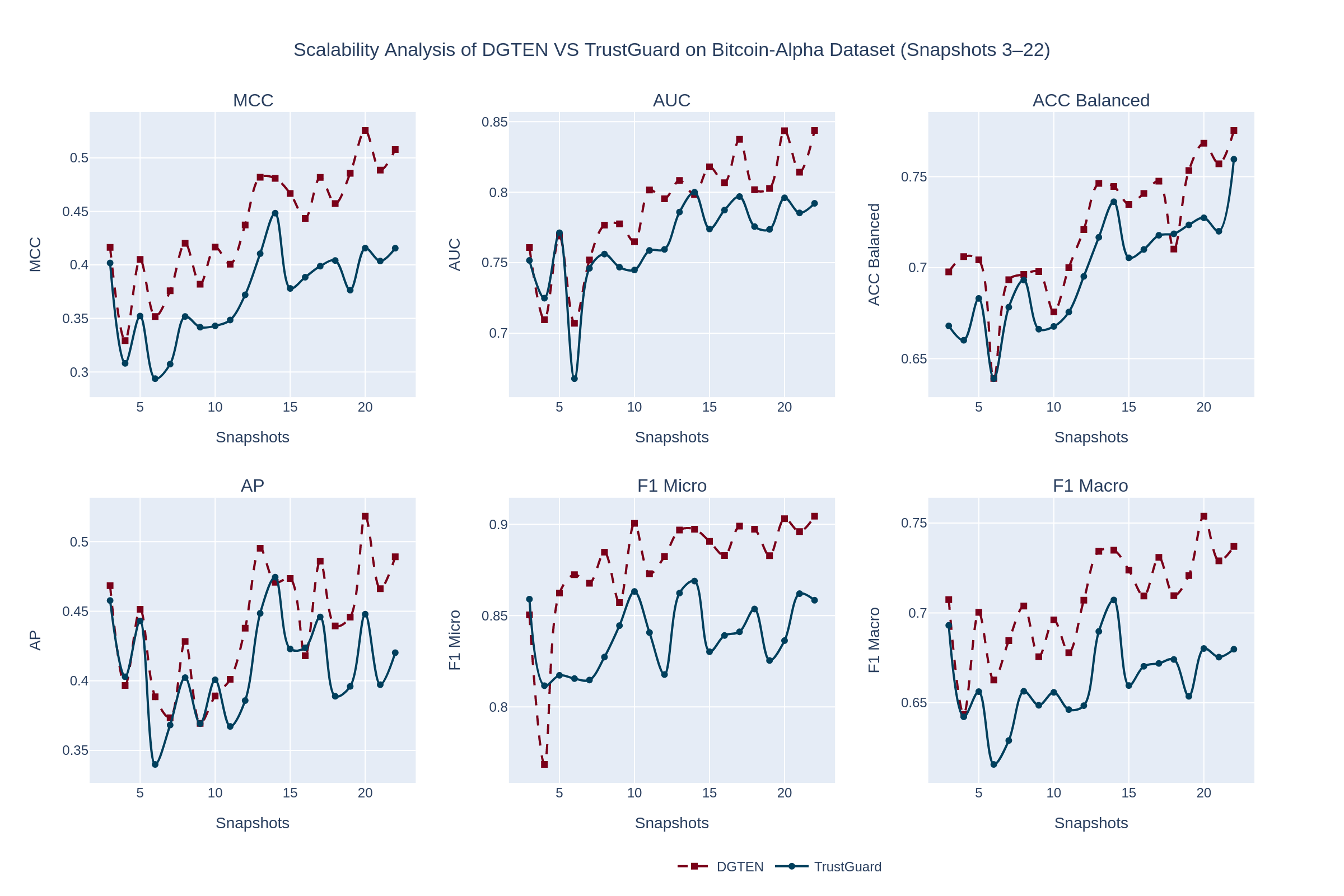}
    \vspace{-0.33mm}
    \caption{\textbf{(Left)} Scalability analysis on Bitcoin-OTC: DGTEN exceeds TrustGuard across 3–19 snapshots (MCC~$\approx 0.50$, AUC mid-0.80s, balanced accuracy~$\approx 0.75$, AP~$\approx 0.51$, F1-micro/macro~$>0.90/0.72$), confirming superior temporal-context utilization for trust prediction, \textbf{(Right)} Scalability analysis on Bitcoin-Alpha (single future-timeslot prediction): DGTEN outperforms TrustGuard across six metrics—MCC, AUC, balanced accuracy, AP, F1-micro, and F1-macro over 3–to-22 snapshots, showing smoother gains with temporal depth and superior use of longer histories for robust trust evaluation.}
    \label{fig:temporal_scalability_alpha}
    \vspace{-0.5em}
\end{figure*}

\subsubsection{DGTEN Scalability Assessment}
We evaluate the scalability of DGTEN across varying temporal depths to assess its practical viability for dynamic trust evaluation. The analysis measures the model’s ability to exploit progressively larger historical contexts—quantified by the number of snapshots—to improve predictive accuracy. Experiments were conducted on the Bitcoin-OTC (Figure~\ref{fig:temporal_scalability_alpha}, 3--19 snapshots) and Bitcoin-Alpha (Figure~\ref{fig:temporal_scalability_alpha}, 3--22 snapshots) datasets, with performance stability and convergence tracked as temporal depth increased.  
TrustGuard serves as the primary baseline as it is the only prior work applying a snapshot-based temporal modeling approach on these datasets. Other models such as Guardian, GATrust, TrustGNN, and KGTrust target static trust networks, making them unsuitable for direct comparison.

On Bitcoin-OTC, DGTEN exhibits a steady rise in performance as snapshots increase from 3 to 19, with MCC improving from approximately 0.30 to 0.4984 at 17 snapshots, a 66\% relative gain. This trend contrasts with TrustGuard’s volatility and earlier peak at 14 snapshots $(MCC \approx 0.4625)$. No performance degradation is observed at maximum depths, indicating robust scalability without overfitting. Improvements are consistent across metrics: MCC, AUC, balanced accuracy, and F1 demonstrating broad performance gains. These results reflect the benefits of DGTEN’s temporal framework, combining HAGH positional encoding, Chebyshev-KAN based attention, and ODE-based refinement.

On Bitcoin-Alpha, DGTEN achieves a peak MCC exceeding 0.52 near 20 snapshots versus TrustGuard’s $\sim 0.45$ peak at 14 snapshots, a 15.6\% advantage. Performance continues improving up to 22 snapshots, showing extended temporal capacity. DGTEN’s curves are smoother and less volatile than TrustGuard’s, particularly in MCC and AUC, underscoring the architecture’s ability to learn effectively from long historical sequences.

Scalability patterns are consistent across datasets despite differences in network density (Bitcoin-OTC: 35{,}592 edges; Bitcoin-Alpha: 24{,}186 edges) and interaction structures, confirming that DGTEN’s temporal mechanisms are domain-independent. An optimal performance window emerges between 17--20 snapshots, beyond which marginal gains diminish, providing guidance for balancing computational cost with accuracy. Quantitatively, DGTEN delivers 7.8\% higher peak MCC on Bitcoin-OTC and 15.6\% on Bitcoin-Alpha compared to TrustGuard, with reduced performance volatility.  

\begin{figure*}[!htbp]
    \centering
    \includegraphics[width=0.99\linewidth]{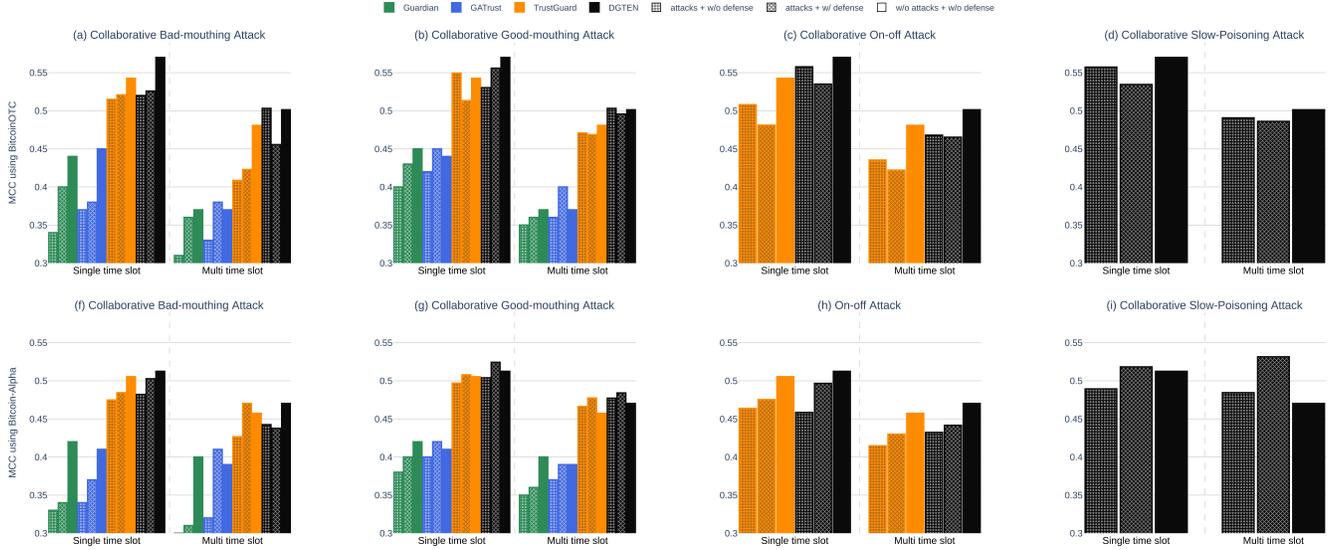}
    \caption{Robustness analysis of trust evaluation methodologies under adversarial conditions. The effectiveness of Guardian, GATrust, TrustGuard, and DGTEN is assessed against three types of collaborative trust-related attacks using the Bitcoin-OTC (a–c) and Bitcoin-Alpha (f–h) datasets, for predicting trust in both single and multiple future time slot settings. We introduced a good mouthing based slow poisoning dynamic attack in (d \& i).}
    \label{fig:attack_analysis}
    \vspace{-3mm}
\end{figure*}

\subsubsection{Adversarial Robustness Analysis}
DGTEN demonstrates strong adversarial resilience on the {Bitcoin-OTC} dataset, maintaining high stability across diverse attack types. Under single-slot collaborative good-mouthing and slow-poisoning attacks, the defense allows the model to retain approximately 94--98\% of its clean performance. Its robustness is even more evident in multi-slot settings: during good-mouthing attacks, DGTEN preserves 99\% of baseline performance despite sustained adversarial interference. More disruptive on-off attacks introduce larger degradation, yet the defended model still recovers to roughly 94\% of its clean accuracy.

On the {Bitcoin-Alpha} dataset, DGTEN exhibits distinct recovery patterns that, in several cases, exceed its clean baseline. With defense enabled, performance surpasses the clean model by 2.3\% in single-slot and 2.9\% in multi-slot good-mouthing attacks. These gains stem from structural differences between the datasets: {Bitcoin-Alpha} is sparser (24{,}186 edges versus 35{,}592 in {Bitcoin-OTC}) and has a higher proportion of trust edges (93.65\% versus 89.99\%). This topology strengthens DGTEN's uncertainty estimation and enables the RAECA mechanism to more effectively isolate and prune adversarial or noisy edges. Consequently, the model achieves strong recovery, improving over the attacked state by up to 8.3\% in single-slot on-off scenarios.

A comparative evaluation against TrustGuard, the strongest Dynamic baseline, further confirms DGTEN's robustness. On {Bitcoin-OTC}, DGTEN consistently exceeds TrustGuard across all comparable attack settings. In single-slot configurations, it achieves gains of approximately 8.3\% under good-mouthing attacks and 11.2\% under on-off attacks. These margins persist in multi-slot scenarios, with improvements of 7.7\% and 10.2\% under bad-mouthing and on-off attacks, respectively. On {Bitcoin-Alpha}, the performance gap is smaller but still systematic: DGTEN outperforms TrustGuard in five of six comparable settings, with notable gains of 3.8\% in single-slot bad-mouthing and 4.5\% in single-slot on-off attacks. The consistency of these improvements across datasets and attack modalities highlights the generality and effectiveness of DGTEN's defense mechanisms.

\begin{table}[h!]
\centering
\caption{Ablation study(mean$\pm$std) on Bitcoin-OTC(Task-1), using 10 snapshots.}
\label{tab:ablation_study}
\resizebox{\columnwidth}{!}{%
\begin{tabular}{lcccc}
\toprule
Model Variant & MCC & AUC & BA & F1-macro \\
Full DGTEN w/ RAECA & 0.437±0.012 & 0.795±0.003 & 0.702±0.004 & 0.715±0.004 \\ 
\midrule
w/o RAECA & 0.431±0.003 & 0.791±0.005 & 0.696±0.003 & 0.710±0.004 \\
w/o ODE   & 0.417±0.003 & 0.775±0.004 & 0.686±0.001 & 0.702±0.001 \\
w/o KAN   & 0.387±0.004 & 0.770±0.002 & 0.691±0.003 & 0.688±0.004 \\
w/ only ODE(takes NEs) & 0.267±0.010 & 0.694±0.007 & 0.641±0.005 & 0.614±0.003 \\
w/o (HAGE,KAN\&ODE) & 0.397±0.003 & 0.771±0.001 & 0.680±0.002 & 0.689±0.001 \\
w/o (KAN\&ODE) & 0.408±0.001 & 0.774±0.000 &0.680±0.006 & 0.693±0.002\\
w/o HAGE & 0.430±0.006 & 0.791±0.003 & 0.703±0.006 & 0.709±0.003\\
\bottomrule
\end{tabular}%
}
\vspace{-3mm}
\end{table}

\subsubsection{Ablation Studies}
To evaluate the individual contributions of the DGTEN framework's components, an ablation study was performed on the Bitcoin-OTC dataset (Task-1). As detailed in Table~\ref{tab:ablation_study}, the Full DGTEN w/ RAECA enabled model achieved superior performance across all metrics (MCC 0.437, AUC 0.795), validating the synergistic integration of the proposed SL and temporal mechanisms.

Within the temporal layer, the Chebyshev-KAN-based multi-headed attention proved critical for feature transformation. Replacing the KAN with a standard linear layer (w/o KAN) resulted in the most significant single-component degradation, reducing MCC by 11.4\% (0.437 to 0.387) and AUC by 3.1\%. This underscores the necessity of high-order non-linear approximations provided by KANs over simple linear projections for capturing complex temporal evolution.

The residual learning-based ODE mechanism also demonstrated substantial contribution; its exclusion (w/o ODE) yielded a 4.6\% decline in MCC. However, the limitation of the ODE acting in isolation was evident in the ``w/ only ODE'' variant, where the NEs from the SL were fed directly into the ODE, bypassing the KAN and HAGE modules. This configuration precipitated a severe 38.9\% decline in MCC. This result indicates that while the ODE is effective for continuous refinement, it cannot compensate for the lack of expressive feature transformation provided by the full temporal stack. Finally, the RAECA module contributed to stability; its removal lowered MCC by 1.4\%, validating the effectiveness of adaptive edge weighting in the final aggregation phase.

\begin{figure*}[!htbp]
    \centering
    \includegraphics[width=0.80\textwidth]{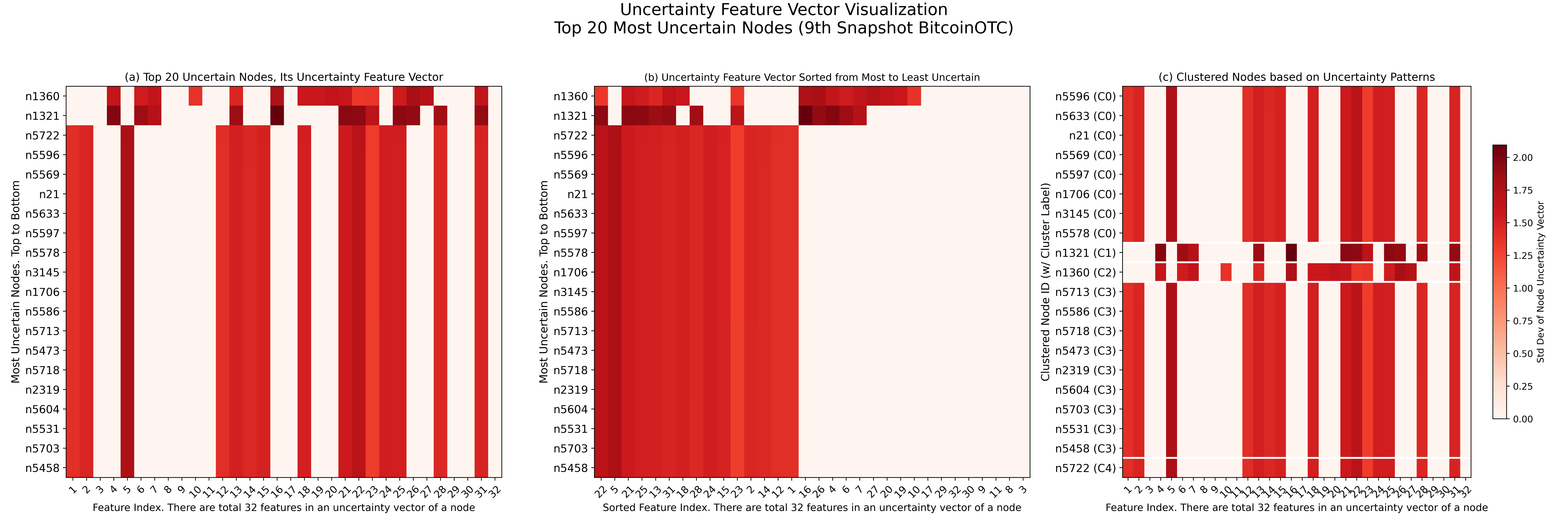}
    \caption{
        Uncertainty Analysis (9th Snapshot, BitcoinOTC).
        (L) Top 20 uncertain nodes with 32-feature fingerprints; darker shades show higher uncertainty. 
        (M) Features sorted by uncertainty, revealing consistent high-risk dimensions. 
        (R) Nodes clustered by pattern similarity, exposing distinct behavioral archetypes.
    }
    \label{fig:static_uncertainty}
    \vspace{-4mm}
\end{figure*}
\begin{figure}[!htbp]
    \centering
    \includegraphics[width=1.\linewidth]{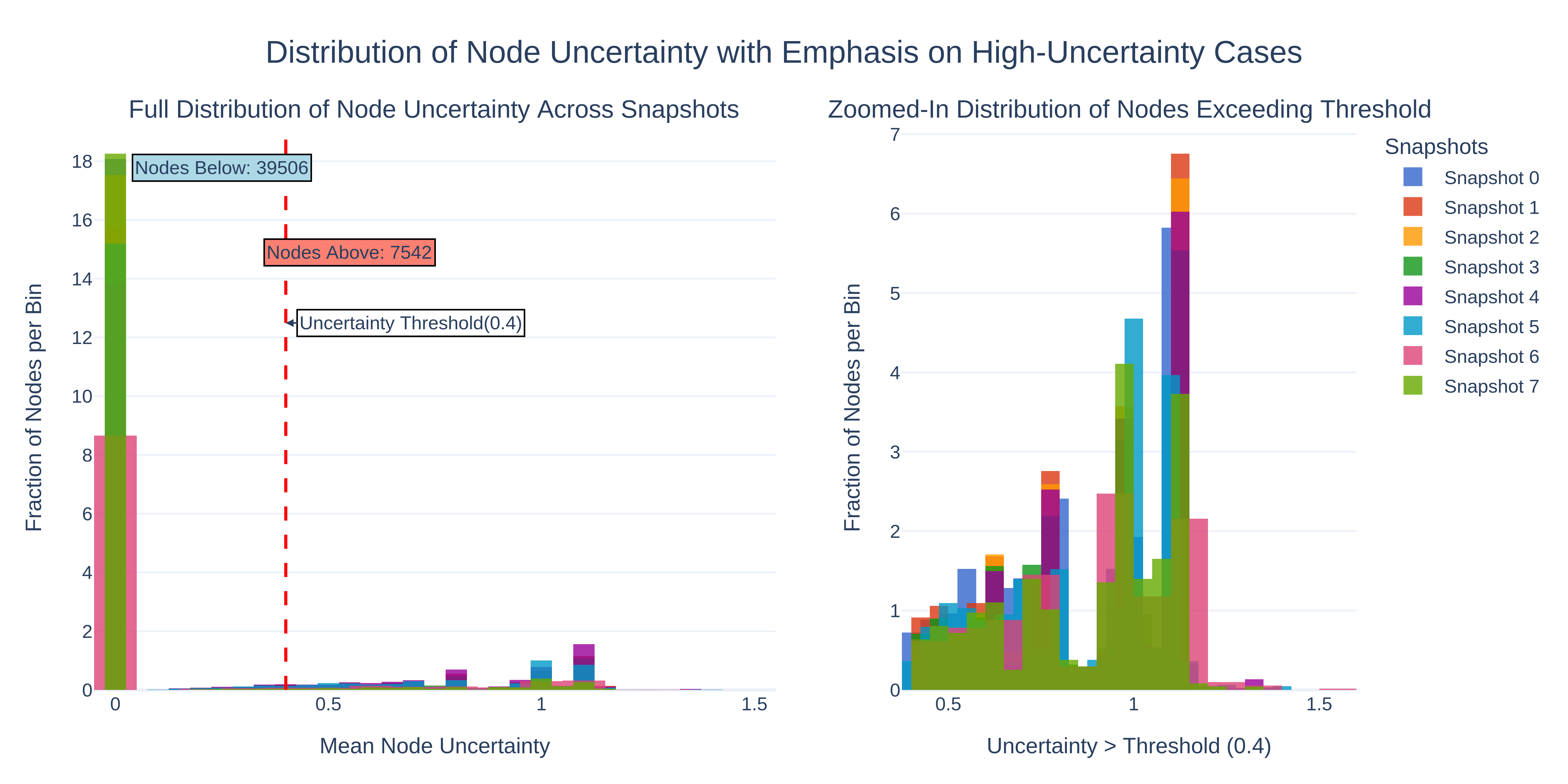}
    \caption{
    Longitudinal Node Uncertainty Analysis.
    \textit{(Left)} Mean node uncertainty with a 0.4 threshold (red dashed) separating stable nodes (39,506) from uncertain nodes (7,542).\textit{(Right)} Zoomed view of nodes above the threshold, showing repeated spikes across Snapshots 3--7 that suggest potential indicators of compromise.
    }
    \label{fig:dynamic_uncertainty}
    \vspace{-4mm}
\end{figure}

\subsubsection{Uncertainty Quantification for Operational Cybersecurity}

DGTEN’s node-level uncertainty provides quantitative diagnostics for both snapshot-level and longitudinal monitoring of dynamic trust graphs. Each snapshot produces a 32-dimensional uncertainty vector for every node through the SL. For snapshot-level inspection, these vectors are analyzed directly to study node uncertainty; for longitudinal monitoring, each vector is reduced to its mean to obtain a scalar uncertainty value per node per snapshot. This separation supports both localized structural analysis and sequence-level tracking of uncertainty evolution. All uncertainty measurements are performed with RAECA disabled.

On the Bitcoin-OTC dataset, uncertainty behavior is examined across ten snapshots. For the snapshot-level analysis (Figure~\ref{fig:static_uncertainty}), the model is trained on snapshots 0–8 and used to predict snapshot 9. We analyze the raw 32-dimensional uncertainty vectors without scalar reduction. Nodes are ranked by their maximum feature-level uncertainty, and the top 20 most uncertain nodes are selected for closer inspection. Sorting and clustering their uncertainty fingerprints exposes snapshot-local structural weaknesses and groups nodes into behavioral archetypes. This highlights nodes whose trust relationships behave inconsistently within the predicted snapshot, allowing us to isolate structurally ambiguous nodes without using longitudinal information.

The longitudinal stage (Figure~\ref{fig:dynamic_uncertainty}) computes, for every snapshot, the mean of each node’s 32-dimensional uncertainty vector, yielding one scalar uncertainty value per node for every snapshot in the sequence. The first two snapshots form the training window, and the third snapshot is treated as the first predicted snapshot (\textit{snapshot-0}) for longitudinal visualization. Tracking these mean uncertainties across the sequence provides a graph-wide stability indicator, showing how uncertainty tightens or disperses as the system processes new trust interactions. Because nodes may appear in multiple snapshots, total counts exceed the number of unique entities; each node–snapshot instance contributes independently to the longitudinal distribution.

A decision threshold on mean uncertainty (e.g., 0.4) provides an operational control point. This threshold is policy-defined rather than learned and determines how conservative automated actions should be. Nodes exceeding the threshold populate a dynamic watchlist, while spikes in above-threshold counts indicate abnormal trust dynamics requiring operator review. This supports risk-aware workflows: low-uncertainty nodes are processed automatically, mid-uncertainty nodes are deferred for verification, and high-uncertainty nodes are withheld from automated trust decisions. These Uncertainty quantification can also enhances threat intelligence and forensics. Historical uncertainty trajectories help localize initial compromise points and reconstruct attack progression.

\vspace{-3mm}
\section{Discussion}
\label{sec:discussion}
The DGTEN model introduced in this work represents a significant advancement in DTE using GNN, particularly in its ability to capture temporal complexities and quantify uncertainty, while remaining robust against trust-related attacks. Experimental results on dynamic graphs consistently demonstrate that DGTEN outperforms existing state-of-the-art methods across multiple dynamic datasets and evaluation tasks.
\vspace{-3mm}
\subsection{Architectural Contributions}
DGTEN’s strength arises from the synergy of three core components. First, the probabilistic representation (DGMP) models each entity in the trust graph as a Gaussian distribution ($\mu,\sigma$), explicitly quantifying uncertainty crucial for cold-start cases with new users, where DGTEN achieves notable gains. This capability extends the model beyond basic prediction to support cybersecurity intelligence applications. Rather than providing binary “trust” or “distrust” outputs, the model reports node-level uncertainty. This enables decision intelligence with tiered operational policies based on confidence. Second, an expressive temporal modeling framework leverages KANs and ODEs to capture complex, evolving trust dynamics. Third, the RAECA robustness module filters malicious or noisy interactions, stabilizing training and reducing performance variance under adversarial conditions.
\subsection{Limitations and Future Directions}
DGTEN inherits several structural constraints. First, the model relies on discretized snapshots, which aggregate all interactions within a time window into a single static graph. This aggregation obscures event ordering and permits high-frequency or compensatory attacks to cancel out within a snapshot, creating blind spots for short-lived or stealthy manipulations.

Second, the RAECA defense is homophily-driven. Its reliance on similarity-based pruning improves robustness on homophilous trust networks but limits generality on heterophilous graphs, leaving the model susceptible to coordinated adversaries capable of engineering artificial similarity.

Third, uncertainty is integrated only at the structural level. Although DGMP produces node-level means and variances for each snapshot, the temporal modules (KAN attention and ODE refinement) are deterministic and can evolve only the expected state (means). They cannot propagate second-order moments (variances) without violating probabilistic consistency. Passing variance vectors through these deterministic pathways would reduce $\sigma$ to ordinary features, resulting in miscalibrated downstream uncertainty. Achieving full uncertainty propagation requires stochastic temporal modules, such as SDE-based dynamics or Bayesian attention mechanisms, which can mathematically evolve distributions rather than point estimates. Consequently, uncertainty currently functions as a structural diagnostic rather than a fully temporally-evolved quantity. Finally, privacy-preserving graph transform methods, such as those proposed by Usman et al.~\cite{usman2025dfdg}, can be adopted as a potential defense against model poisoning attacks. 

These limitations outline clear paths for future research. From a scope perspective, the present research work is intentionally focused on advancing the architectural foundations of dynamic trust evaluation with node-level uncertainty quantification. Introducing DGMP, ODE-based residual learning, and $2^{nd}$ degree Chebyshev polynomial based KAN-based multi-head attention within a single unified framework constitutes a substantial methodological contribution. Addressing generalizing beyond homophily assumptions, protecting model poisoning, and developing stochastic temporal architectures are separate methodological problems that naturally define the next steps for this line of work.

\section{Conclusion}
\label{sec:conclusion}
In this paper, we introduced DGTEN, a novel Deep Gaussian-based GNN framework designed to address critical gaps in dynamic trust evaluation for cybersecurity applications. By integrating uncertainty-aware message passing, advanced temporal modeling through HAGH positional encoding, Chebyshev-KAN for multi-head attention, and neural ODE residual learning, alongside the RAECA for adversarial defense, DGTEN provides a comprehensive solution for modeling evolving trust dynamics in complex networked systems. Our evaluations on the Bitcoin-OTC and Bitcoin-Alpha datasets demonstrate DGTEN's superior performance, achieving MCC gains of up to 12.34\% for single-timeslot prediction, 10.07\% for multi-timeslot forecasting, and 25.00\% in cold-start scenarios, together with AUC, balanced accuracy, and F1-macro improvements of up to 4.48\%, 8.72\%, and 6.93\%, respectively, over strong baselines such as TrustGuard, Guardian, and GATrust.
DGTEN also exhibits consistent adversarial robustness, retaining up to 98--99\% of clean performance under collaborative attacks, surpassing baseline accuracy by as much as 2.9\% in defended settings, and outperforming TrustGuard by margins reaching 11.2\% depending on the attack type.

These results underscore DGTEN's ability to capture temporal evolution, quantify epistemic uncertainty, and maintain robustness against sophisticated attacks such as bad-mouthing, good-mouthing, and on-off behaviors. The framework's uncertainty quantification further enables actionable cybersecurity insights, such as identifying high-risk nodes and behavioral anomalies, facilitating risk-aware decision-making in real-world systems like IoT networks, social platforms, and financial ecosystems.

While DGTEN advances the field, opportunities for enhancement remain, including hybrid continuous-time modeling to overcome snapshot aggregation limitations, entire-model poisoning attacks and deeper integration of uncertainty into predictive logic for fully adaptive systems. Future work will explore these extensions, alongside applications to heterophilous graphs and broader domains, to further strengthen trust evaluation in dynamic, adversarial environments.

\vspace{-1mm}
\section{Conflict of interest}
The authors declare that there are no conflicts of interest regarding the publication of this paper.
\vspace{-2mm}
\section{Source Code}
The authors will release the source code upon publication, including two separate projects covering static and dynamic modeling.

\bibliographystyle{IEEEtran}
\bibliography{DGTEN-Paper}

@String{Computing = "Computing" }

@String{Computer = "{IEEE} Computer" }

@article{wang2024trustguard,
  title={Trustguard: Gnn-based robust and explainable trust evaluation with dynamicity support},
  author={Wang, Jie and Yan, Zheng and Lan, Jiahe and Bertino, Elisa and Pedrycz, Witold},
  journal={IEEE Transactions on Dependable and Secure Computing},
  year={2024},
  publisher={IEEE}
}

@article{chen2016trust,
  title={Trust-based service management for social internet of things systems},
  author={Chen, Ing-Ray and Bao, Feng and Guo, Jin},
  journal={IEEE Transactions on Dependable and Secure Computing},
  volume={13},
  number={6},
  pages={684--696},
  year={2016},
  publisher={IEEE}
}

@article{chen2014trust,
  title={Trust management for SOA-based IoT and its application to service composition},
  author={Chen, Ray and Guo, Jia and Bao, Fenye},
  journal={IEEE Transactions on Services Computing},
  volume={9},
  number={3},
  pages={482--495},
  year={2014},
  publisher={IEEE}
}

@article{wang2020survey,
  title={A survey on trust evaluation based on machine learning},
  author={Wang, Jingwen and Jing, Xuyang and Yan, Zheng and Fu, Yulong and Pedrycz, Witold and Yang, Laurence T},
  journal={ACM Computing Surveys (CSUR)},
  volume={53},
  number={5},
  pages={1--36},
  year={2020},
  publisher={ACM New York, NY, USA}
}

@article{yan2014survey,
  title={A survey on trust management for Internet of Things},
  author={Yan, Zheng and Zhang, Peng and Vasilakos, Athanasios V},
  journal={Journal of network and computer applications},
  volume={42},
  pages={120--134},
  year={2014},
  publisher={Elsevier}
}

@inproceedings{liu2019neuralwalk,
  title={Neuralwalk: Trust assessment in online social networks with neural networks},
  author={Liu, Guanglai and Li, Chaoyi and Yang, Qiang},
  booktitle={IEEE INFOCOM 2019-IEEE Conference on Computer Communications},
  pages={1999--2007},
  year={2019},
  organization={IEEE}
}

@article{huo2023trustgnn,
  title={Trustgnn: Graph neural network-based trust evaluation via learnable propagative and composable nature},
  author={Huo, Chen and He, Di and Liang, Chuan and Jin, Dong and Qiu, Tong and Wu, Liang},
  journal={IEEE Transactions on Neural Networks and Learning Systems},
  year={2023},
  publisher={IEEE}
}

@inproceedings{yu2023kgtrust,
  title={Kgtrust: Evaluating trustworthiness of siot via knowledge enhanced graph neural networks},
  author={Yu, Ziyu and Jin, Dong and Huo, Chen and Wang, Zhan and Liu, Xiao and Qi, Honggang and Wu, Jie and Wu, Liang},
  booktitle={Proceedings of the ACM Web Conference 2023},
  pages={727--736},
  year={2023},
  organization={ACM}
}

@article{li2020guardian,
  title={Guardian: Evaluating trust in online social networks with graph convolutional networks},
  author={Li, Wei and Gao, Zheng and Li, Bo},
  journal={IEEE INFOCOM 2020-IEEE Conference on Computer Communications},
  pages={914--923},
  year={2020},
  publisher={IEEE}
}

@article{gao2020gatrust,
  title={GATrust: Leveraging multi-aspect properties for trust evaluation with graph attention networks},
  author={Gao, Zheng and Li, Wei and Liu, Bing},
  journal={IEEE INFOCOM 2020-IEEE Conference on Computer Communications},
  pages={914--923},
  year={2020},
  publisher={IEEE}
}

@article{lin2021medley,
  title={Medley: Predicting social trust in time-varying online social networks},
  author={Lin, Wei and Li, Bo},
  journal={IEEE INFOCOM 2021-IEEE Conference on Computer Communications},
  pages={1--10},
  year={2021},
  publisher={IEEE}
}

@inproceedings{grover2016node2vec,
  title={node2vec: Scalable feature learning for networks},
  author={Grover, Aditya and Leskovec, Jure},
  booktitle={Proceedings of the 22nd ACM SIGKDD International Conference on Knowledge Discovery and Data Mining},
  pages={855--864},
  year={2016}
}

@inproceedings{jagielski2021subpopulation,
  title={Subpopulation data poisoning attacks},
  author={Jagielski, Matthew and Severi, Giorgio and Pousette Harger, Niklas and Oprea, Alina},
  booktitle={Proceedings of the 2021 ACM SIGSAC Conference on Computer and Communications Security},
  pages={3104--3122},
  year={2021}
}

@INPROCEEDINGS{7837846,
  author={Kumar, Srijan and Spezzano, Francesca and Subrahmanian, V. S. and Faloutsos, Christos},
  booktitle={2016 IEEE 16th International Conference on Data Mining (ICDM)}, 
  title={Edge Weight Prediction in Weighted Signed Networks}, 
  year={2016},
  volume={},
  number={},
  pages={221-230},
}

@ARTICLE{10925363,
  author={Luo, Tingxi and Wang, Jie and Yan, Zheng and Gelenbe, Erol},
  journal={IEEE Network}, 
  title={Graph Neural Networks for Trust Evaluation: Criteria, State-of-the-Art, and Future Directions}, 
  year={2025},
  volume={},
  number={},
  pages={1-1}
}

@ARTICLE{9936655,
  author={Yin, Xiaoyan and Lin, Wanyu and Sun, Kexin and Wei, Chun and Chen, Yanjiao},
  journal={IEEE Transactions on Information Forensics and Security}, 
  title={A2S2-GNN: Rigging GNN-Based Social Status by Adversarial Attacks in Signed Social Networks}, 
  year={2023},
  volume={18},
  number={},
  pages={206-220},
}

@inproceedings{ijcai2019p669,
  title     = {Adversarial Examples for Graph Data: Deep Insights into Attack and Defense},
  author    = {Wu, Huijun and Wang, Chen and Tyshetskiy, Yuriy and Docherty, Andrew and Lu, Kai and Zhu, Liming},
  booktitle = {Proceedings of the Twenty-Eighth International Joint Conference on
               Artificial Intelligence, {IJCAI-19}},
  pages     = {4816--4823},
  year      = {2019},
  month     = {7},
}

@inproceedings{10.1145/3394486.3403049,
author = {Jin, Wei and Ma, Yao and Liu, Xiaorui and Tang, Xianfeng and Wang, Suhang and Tang, Jiliang},
title = {Graph Structure Learning for Robust Graph Neural Networks},
year = {2020},
isbn = {9781450379984},
publisher = {Association for Computing Machinery},
address = {New York, NY, USA},
booktitle = {Proceedings of the 26th ACM SIGKDD International Conference on Knowledge Discovery \& Data Mining},
pages = {66–74},
numpages = {9},
location = {Virtual Event, CA, USA},
series = {KDD '20}
}

@ARTICLE{9546070,
  author={Wang, Qi and Zhao, Weiliang and Yang, Jian and Wu, Jia and Xue, Shan and Xing, Qianli and Yu, Philip S.},
  journal={IEEE Transactions on Neural Networks and Learning Systems}, 
  title={C-DeepTrust: A Context-Aware Deep Trust Prediction Model in Online Social Networks}, 
  year={2023},
  volume={34},
  number={6},
  pages={2767-2780},
}

@article{facd5bfb-620b-3f71-8412-0c6d4aa71a2e,
 ISSN = {03600572, 15452115},
 author = {Miller McPherson and Lynn Smith-Lovin and James M. Cook},
 journal = {Annual Review of Sociology},
 pages = {415--444},
 publisher = {Annual Reviews},
 title = {Birds of a Feather: Homophily in Social Networks},
 urldate = {2025-06-10},
 volume = {27},
 year = {2001}
}

@inproceedings{10.1145/2433396.2433405,
author = {Tang, Jiliang and Gao, Huiji and Hu, Xia and Liu, Huan},
title = {Exploiting homophily effect for trust prediction},
year = {2013},
isbn = {9781450318693},
publisher = {Association for Computing Machinery},
address = {New York, NY, USA},
booktitle = {Proceedings of the Sixth ACM International Conference on Web Search and Data Mining},
pages = {53–62},
numpages = {10},
location = {Rome, Italy},
series = {WSDM '13}
}

@article{vaswani2017attention,
  title={Attention is all you need},
  author={Vaswani, Ashish and Shazeer, Noam and Parmar, Niki and Uszkoreit, Jakob and Jones, Llion and Gomez, Aidan N and Kaiser, {\L}ukasz and Polosukhin, Illia},
  journal={Advances in neural information processing systems},
  volume={30},
  year={2017}
}

@article{zhu2020beyond,
  title={Beyond homophily in graph neural networks: Current limitations and effective designs},
  author={Zhu, Jiong and Yan, Yujun and Zhao, Lingxiao and Heimann, Mark and Akoglu, Leman and Koutra, Danai},
  journal={Advances in neural information processing systems},
  volume={33},
  pages={7793--7804},
  year={2020}
}

@article{Jin_Feng_Guo_Wang_Wei_Wang_2023, 
title={Local-Global Defense against Unsupervised Adversarial Attacks on Graphs}, 
volume={37}, 
number={7}, 
journal={Proceedings of the AAAI Conference on Artificial Intelligence}, 
author={Jin, Di and Feng, Bingdao and Guo, Siqi and Wang, Xiaobao and Wei, Jianguo and Wang, Zhen}, year={2023}, 
month={Jun.}, 
pages={8105-8113} }

@article{10.1109/TKDE.2022.3201243,
author = {Sun, Lichao and Dou, Yingtong and Yang, Carl and Zhang, Kai and Wang, Ji and Yu, Philip S. and He, Lifang and Li, Bo},
title = {Adversarial Attack and Defense on Graph Data: A Survey},
year = {2023},
issue_date = {Aug. 2023},
publisher = {IEEE Educational Activities Department},
address = {USA},
volume = {35},
number = {8},
issn = {1041-4347},
journal = {IEEE Trans. on Knowl. and Data Eng.},
month = aug,
pages = {7693–7711},
numpages = {19}
}

@inproceedings{10.5555/3294771.3294869,
author = {Hamilton, William L. and Ying, Rex and Leskovec, Jure},
title = {Inductive representation learning on large graphs},
year = {2017},
isbn = {9781510860964},
publisher = {Curran Associates Inc.},
address = {Red Hook, NY, USA},
booktitle = {Proceedings of the 31st International Conference on Neural Information Processing Systems},
pages = {1025–1035},
numpages = {11},
location = {Long Beach, California, USA},
series = {NIPS'17}
}

@article{defazio2022momentumized,
  title={A momentumized, adaptive, dual averaged gradient method},
  author={Defazio, Aaron and Jelassi, Samy},
  journal={Journal of Machine Learning Research},
  volume={23},
  number={144},
  pages={1--34},
  year={2022}
}

@inproceedings{10.1145/3460120.3484805,
author = {Zhang, Yihe and Yuan, Xu and Li, Jin and Lou, Jiadong and Chen, Li and Tzeng, Nian-Feng},
title = {Reverse Attack: Black-box Attacks on Collaborative Recommendation},
year = {2021},
isbn = {9781450384544},
publisher = {Association for Computing Machinery},
address = {New York, NY, USA},
booktitle = {Proceedings of the 2021 ACM SIGSAC Conference on Computer and Communications Security},
pages = {51–68},
numpages = {18},
location = {Virtual Event, Republic of Korea},
series = {CCS '21}
}

@article{JAFARIAN2025125391,
title = {Using attentive temporal GNN for dynamic trust assessment in the presence of malicious entities},
journal = {Expert Systems with Applications},
volume = {260},
pages = {125391},
year = {2025},
issn = {0957-4174},
author = {Besat Jafarian and Nasser Yazdani and Mohammad {Sayad Haghighi}},
}

@article{suarez2022graph,
  title={Graph neural networks for communication networks: Context, use cases and opportunities},
  author={Su{\'a}rez-Varela, Jos{\'e} and Almasan, Paul and Ferriol-Galm{\'e}s, Miquel and Rusek, Krzysztof and Geyer, Fabien and Cheng, Xiangle and Shi, Xiang and Xiao, Shihan and Scarselli, Franco and Cabellos-Aparicio, Albert and others},
  journal={IEEE network},
  volume={37},
  number={3},
  pages={146--153},
  year={2022},
  publisher={IEEE}
}

@article{li2024permutation,
  title={Permutation equivariant graph framelets for heterophilous graph learning},
  author={Li, Jianfei and Zheng, Ruigang and Feng, Han and Li, Ming and Zhuang, Xiaosheng},
  journal={IEEE Transactions on neural networks and learning systems},
  volume={35},
  number={9},
  pages={11634--11648},
  year={2024},
  publisher={IEEE}
}

@inproceedings{wen2023dtrust,
  title={Dtrust: Toward dynamic trust levels assessment in time-varying online social networks},
  author={Wen, Jie and Jiang, Nan and Li, Jin and Liu, Ximeng and Chen, Honglong and Ren, Yanzhi and Yuan, Zhaohui and Tu, Ziang},
  booktitle={IEEE INFOCOM 2023-IEEE Conference on Computer Communications},
  pages={1--10},
  year={2023},
  organization={IEEE}
}

@article{zhan2024enhancing,
  title={Enhancing worker recruitment in collaborative mobile crowdsourcing: A graph neural network trust evaluation approach},
  author={Zhan, Zhongwei and Wang, Yingjie and Duan, Peiyong and Sai, Akshita Maradapu Vera Venkata and Liu, Zhaowei and Xiang, Chaocan and Tong, Xiangrong and Wang, Weilong and Cai, Zhipeng},
  journal={IEEE Transactions on Mobile Computing},
  volume={23},
  number={10},
  pages={10093--10110},
  year={2024},
  publisher={IEEE}
}

@article{jiang2024tfd,
  title={Tfd: Trust-based fraud detection in siot with graph convolutional networks},
  author={Jiang, Nan and Gu, Weihao and Li, Lang and Zhou, Fengqi and Qiu, Sen and Zhou, Tianqing and Chen, Honglong},
  journal={IEEE Transactions on Consumer Electronics},
  volume={71},
  number={1},
  pages={1897--1908},
  year={2024},
  publisher={IEEE}
}

@article{wang2024joint,
  title={Joint item recommendation and trust prediction with graph neural networks},
  author={Wang, Gang and Wang, Hanru and Gong, Junqiao and Ma, Jingling},
  journal={Knowledge-Based Systems},
  volume={285},
  pages={111340},
  year={2024},
  publisher={Elsevier}
}

@inproceedings{bellaj2023gbtrust,
  title={Gbtrust: Leveraging edge attention in graph neural networks for trust management in p2p networks},
  author={Bellaj, Badr and Ouaddah, Aafaf and Mezrioui, Abdelattif and Crespi, Noel and Bertin, Emmanuel},
  booktitle={2023 IEEE 22nd International Conference on Trust, Security and Privacy in Computing and Communications (TrustCom)},
  pages={1272--1278},
  year={2023},
  organization={IEEE}
}

@article{usman2025dfdg,
  title={DFDG: Adaptive federated learning for dynamic graph-based traffic forecasting},
  author={Usman, Muhammad and Lee, Yugyung},
  journal={Knowledge-Based Systems},
  pages={114019},
  year={2025},
  publisher={Elsevier}
}

@ARTICLE{4700287,
  author={Scarselli, Franco and Gori, Marco and Tsoi, Ah Chung and Hagenbuchner, Markus and Monfardini, Gabriele},
  journal={IEEE Transactions on Neural Networks}, 
  title={The Graph Neural Network Model}, 
  year={2009},
  volume={20},
  number={1},
  pages={61-80},
  doi={10.1109/TNN.2008.2005605}}

@inproceedings{massa2009bowling,
  title={Bowling alone and trust decline in social network sites},
  author={Massa, Paolo and Salvetti, Martino and Tomasoni, Danilo},
  booktitle={2009 Eighth IEEE international conference on dependable, autonomic and secure computing},
  pages={658--663},
  year={2009},
  organization={IEEE}
}

@inproceedings{rossi2015network,
  title={The network data repository with interactive graph analytics and visualization},
  author={Rossi, Ryan and Ahmed, Nesreen},
  booktitle={Proceedings of the AAAI conference on artificial intelligence},
  volume={29},
  number={1},
  year={2015}
}

@inproceedings{liu2017opinionwalk,
  title={Opinionwalk: An efficient solution to massive trust assessment in online social networks},
  author={Liu, Guangchi and Chen, Qi and Yang, Qing and Zhu, Binhai and Wang, Honggang and Wang, Wei},
  booktitle={IEEE INFOCOM 2017-IEEE Conference on Computer Communications},
  pages={1--9},
  year={2017},
  organization={IEEE}
}

@inproceedings{yao2013matri,
  title={Matri: a multi-aspect and transitive trust inference model},
  author={Yao, Yuan and Tong, Hanghang and Yan, Xifeng and Xu, Feng and Lu, Jian},
  booktitle={Proceedings of the 22nd international conference on World Wide Web},
  pages={1467--1476},
  year={2013}
}

@inproceedings{liu2014assessment,
  title={Assessment of multi-hop interpersonal trust in social networks by three-valued subjective logic},
  author={Liu, Guangchi and Yang, Qing and Wang, Honggang and Lin, Xiaodong and Wittie, Mike P},
  booktitle={IEEE INFOCOM 2014-IEEE Conference on Computer Communications},
  pages={1698--1706},
  year={2014},
  organization={IEEE}
}

@inproceedings{massa2005controversial,
  title={Controversial users demand local trust metrics: An experimental study on epinions. com community},
  author={Massa, Paolo and Avesani, Paolo},
  booktitle={AAAI},
  volume={1},
  pages={121--126},
  year={2005}
}

\end{document}